%% file: main.tex
\documentclass[11pt]{article}

\usepackage[preprint]{acl}

\usepackage{times}
\usepackage{latexsym}

\usepackage[T1]{fontenc}

\usepackage[utf8]{inputenc}

\usepackage{microtype}

\usepackage{inconsolata}

\usepackage{graphicx}

\usepackage{titletoc}
\newcommand\DoToC{%
  \startcontents
  \printcontents{}{1}{}{}
}

\newcommand{\squishlist}{
\begin{list}{$\bullet$}
{   \setlength{\itemsep}{0pt}
   \setlength{\parsep}{3pt}
   \setlength{\topsep}{3pt}
   \setlength{\partopsep}{0pt}
   \setlength{\leftmargin}{1.5em}
   \setlength{\labelwidth}{1em}
   \setlength{\labelsep}{0.5em} } }
\newcounter{Lcount}
\newcommand{\squishlisttwo}{
\begin{list}{\arabic{Lcount}. }
  { \usecounter{Lcount}
 \setlength{\itemsep}{0pt}
 \setlength{\parsep}{0pt}
 \setlength{\topsep}{0pt}
 \setlength{\partopsep}{0pt}
 \setlength{\leftmargin}{1.5em}
 \setlength{\labelwidth}{1em}
 \setlength{\labelsep}{0.5em} } }
\newcommand{\squishend}{\end{list} }
\usepackage{enumitem}
\usepackage{csquotes}
\usepackage{makecell}

\usepackage{graphicx}
\usepackage{amsmath}
\usepackage{amssymb}
\usepackage{dsfont}
\usepackage{booktabs}
\usepackage{multirow}
\usepackage{array}
\usepackage{subcaption}
\usepackage{longtable}
\usepackage{bbm}
\usepackage{adjustbox}
\usepackage{hyperref}
\usepackage{cleveref}
\usepackage{algorithm}
\usepackage{algpseudocode}

\usepackage{tabularx}
\usepackage{tcolorbox}
\tcbuselibrary{skins}
\usepackage[dvipsnames]{xcolor}

\title{Belief Coevolution in a Social Network of Generalist and Specialist Large Language Models: A Controlled Study}

\author{
 \textbf{Germans Savcisens\textsuperscript{1,2}},
 \textbf{Samantha Dies\textsuperscript{1}},
 \textbf{Courtney Maynard\textsuperscript{1}},
 \textbf{Tina Eliassi-Rad\textsuperscript{1,2,3}}
\\
 \textsuperscript{1}Khoury College of Computer Sciences, Northeastern University, Boston, USA\\
 \textsuperscript{2}Network Science Institute, Northeastern University, Boston, USA\\
 \textsuperscript{3}Santa Fe Institute, Santa Fe, USA
\\
 \small{
   \textbf{Correspondence:} \href{mailto:germans@savcisens.com}{germans@savcisens.com}
 }
}

\hypersetup{
  pdftitle={Belief Coevolution in a Social Network of Generalist and Specialist Large Language Models: A Controlled Study},
  pdfauthor={Germans Savcisens and Samantha Dies and Courtney Maynard and Tina Eliassi-Rad},
  pdfsubject={Belief diffusion in multi-agent LLM populations},
  pdfkeywords={multi-agent LLMs, opinion dynamics, belief diffusion, mixed-effects models, network science}
}

\begin{document}
\maketitle

\begin{abstract}
Large language models (LLMs) are increasingly deployed in multi-agent environments. However, the processes by which beliefs form and propagate among interacting LLMs remain poorly understood.
We introduce \texttt{CoevolveSim}, a framework for studying belief diffusion within networked LLM populations.  
\texttt{CoevolveSim} allows us to isolate and study three factors: domain specialization, social-role assignment, and social network structure.
Within this framework, generalist and specialist LLM agents exchange and revise beliefs. In each round, an LLM agent observes a summary of its neighbors' beliefs before updating its own.
We find that persona-style role assignment reshapes individual belief revision but has little effect on population-level consensus. 
In contrast, introducing finetuned specialist LLMs produces a large change in consensus and is accompanied by persistent asymmetries in outgoing influence. 
We further show that an opinion-dynamics surrogate combining individual persistence with population-level belief composition reproduces consensus with high fidelity in every scenario, whereas predicting individual final-round beliefs in heterogeneous LLM populations additionally benefits from agent identity. 
Code and simulation data: 
\href{https://github.com/carlomarxdk/coevolve-sim}{github.com/carlomarxdk/coevolve-sim}.
\end{abstract}

\input{sections/1_introduction}

\input{sections/2_related_works}

\input{sections/3_methods}

\input{sections/4_setup}

\input{sections/5_results}

\input{sections/6_discussion}

\input{sections/7_conclusion}

\bibliography{refs-condensed}

\appendix
\setcounter{figure}{0}
\setcounter{table}{0}
\setcounter{equation}{0}
\renewcommand{\thefigure}{A\arabic{figure}}
\renewcommand{\thetable}{A\arabic{table}}
\renewcommand{\theequation}{A\arabic{equation}}

\clearpage
\onecolumn
\section*{Appendix}
\DoToC

\clearpage
\input{sections/appendix/A_notations}
\clearpage
\input{sections/appendix/B_taxonomy}
\clearpage
\input{sections/appendix/C_LLMs}
\clearpage
\input{sections/appendix/D_maximin}
\clearpage
\input{sections/appendix/F_influence}
\clearpage
\input{sections/appendix/E_descriptive_analysis}

\clearpage
\input{sections/appendix/G_calibration_check}
\clearpage
\input{sections/appendix/H_surrogate_models}
\clearpage
\input{sections/appendix/Y_tables}

\end{document}

%% file: sections/1_introduction.tex
\section{Introduction}
\label{sec:introduction}
\begin{figure*}[ht]
\centering
\includegraphics[width=0.95\textwidth]{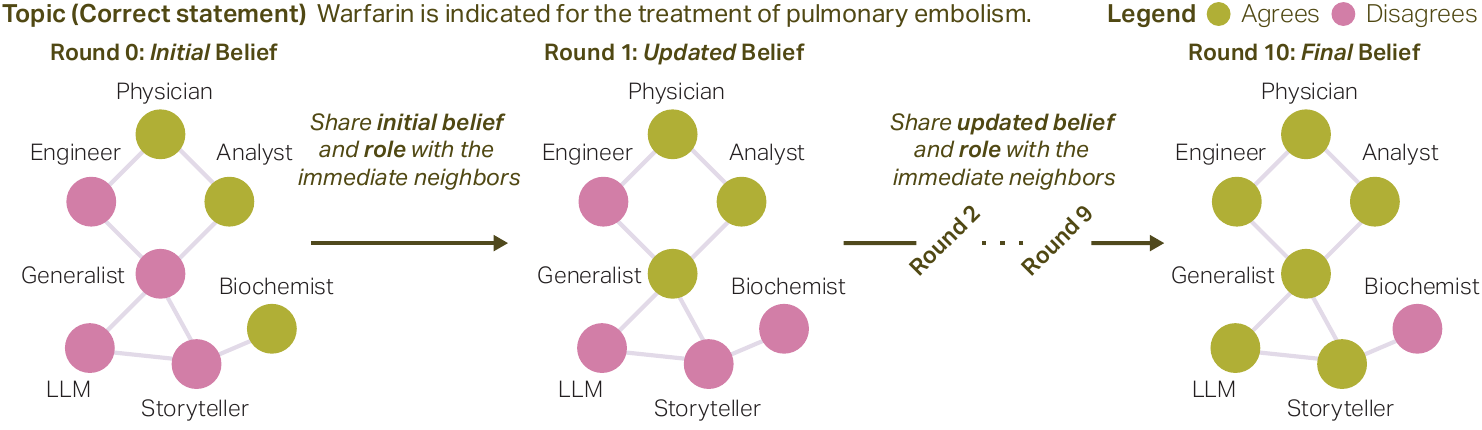}
\caption{\textbf{Schematic overview of \texttt{CoevolveSim} belief diffusion.} Given a discussion topic, we first ask each agent for its initial belief without providing information about the neighbors (Round \(0\)). Then, \texttt{CoevolveSim} facilitates the exchange of beliefs and social roles between agents and their immediate neighbors. In each subsequent round, we evaluate the beliefs of each agent after providing a summary of the beliefs of its neighbors from the previous round. After \(T=10\) rounds of interaction, agents reach their final belief states.}
\label{fig:flow}
\end{figure*}

Text generated by large language models (LLMs) appears in contexts ranging from UN press releases to scientific articles~\cite{liang2025widespread,liang2025quantifying}. On \texttt{Quora}, the share of AI-generated posts rose from 2\% in 2022 to 39\% by late 2024~\cite{sun-etal-2025-ai}.
At the same time, LLMs are increasingly common in multi-agent settings, including debate and collaborative reasoning~\cite{guo2024large,wu_autogen_2024,tran2025multi}. As these applications grow, understanding how beliefs form and spread within populations of interacting LLMs becomes increasingly important.

Individuals revise beliefs after exposure to others' opinions~\cite{lorenz2011social} and often reach consensus after repeated interactions~\cite{becker2017network}.
Nevertheless, the degree to which individuals influence others' beliefs varies unevenly within a population~\cite{moussaid2013social} and depends on the social status of individuals~\cite{bailey2023meta}.
Hence, collective outcomes emerge from an \textit{interplay} between individual characteristics and social structure rather than from any single mechanism in isolation.
Existing work on multi-agent LLM systems has largely focused on behavioral outcomes such as debate performance~\cite{guo2024large}.
However, LLM-based agents may exhibit varied behaviors as they are powered by different LLMs, assigned certain social roles or connected to distinct sets of agents in the interaction networks.
\citet{ng2026mass} and \citet{ye2026stop} highlight that the relationship between these factors remains largely understudied, making it difficult to quantify their contributions to belief dynamics. 

We introduce \texttt{CoevolveSim}, a network-mediated belief diffusion framework in which generalist and specialist LLM agents exchange and revise beliefs over repeated rounds of interaction (see \Cref{fig:flow}). 
In \texttt{CoevolveSim}, we focus on three factors:
(1) \textit{Domain specialization}, which specifies the finetuned LLM that powers an agent (e.g., a model specialized in medicine, chemistry, or mathematics). (2) \textit{Social role}, which is assigned to each LLM agent. This persona-style label is communicated to the agent and its neighbors. (3) \textit{Structure of the social network}, which determines pairwise connections between LLM agents. These connections allow LLM agents to exchange beliefs. This setup allows us to answer two questions, which we study across four experimental scenarios~(\S\ref{sec:simulations}): (1) Which factors shape belief diffusion in populations of interacting LLM agents? (2) Can the resulting dynamics be explained using classical opinion-dynamics mechanisms?
We quantify the resulting belief dynamics in terms of belief revision, exerted influence, and consensus (\S\ref{sec:descriptive}). 
We then assess whether opinion-dynamics models (referred to here as surrogate models) can reproduce these dynamics by predicting belief updates from an agent’s current belief, global or local belief composition, and agent identity (\S\ref{sec:predictive}).

Our \textbf{contributions} are as follows: 
\squishlisttwo
    \item We introduce \texttt{CoevolveSim}, a framework for studying belief diffusion in populations of networked LLM agents; 
    \item Using \(2{,}560\) controlled simulations across four scenarios, we disentangle domain specialization, social-role assignment, and network structure and show that role assignment reshapes individual belief trajectories yet leaves consensus nearly unchanged (\S\ref{sec:descriptive_results}); 
    \item We show that LLM heterogeneity accounts for most of collective belief change, giving rise to opinion-leader-like influence asymmetries;
    \item We show that the mechanisms predicting individual belief revisions differ from those that reproduce collective consensus~(\S\ref{sec:surrogate_results}).
\squishend

%% file: sections/2_related_works.tex
\section{Related Work}
\label{sec:background}
Our work stands at the intersection of research on multi-agent LLM systems, heterogeneous agent populations, and opinion dynamics.

\textbf{Beliefs in multi-agent LLM systems.} Multi-agent LLM systems are typically evaluated through behavioral outcomes such as task completion, voting, and debate performance~\cite{guo2024large,wu_autogen_2024,chen_agentverse_2024}.
More recent work focuses directly on the belief states, showing that belief consistency varies across tasks and models~\cite{pal_large_2025}.
Similarly, population-level outcomes such as consensus can also amplify persuasion~\cite{bilgin_effect_2025}, further highlighting that interactions between \textit{multiple} factors affect collective behavior~\cite{brockers_disentangling_2026}. 
However, agent states, local interactions, and population-level dynamics remain jointly understudied~\cite{ng2026mass}, and existing work rarely isolates the effects of domain specialization, social roles, and network structure.
Refer to the detailed taxonomy in~\Cref{app:multiagenttaxonomy}.

\textbf{Domain specialization and social roles.}
Heterogeneity in multi-agent LLM populations is commonly introduced through either prompt-based personas (that instruct an LLM on how to behave) or specialized models (that use various pretrained or finetuned LLMs). 
Persona-based approaches simulate diverse social identities~\cite{chuang-etal-2024-simulating,taubenfeld-etal-2024-systematic}, but their effects often weaken over extended interactions~\cite{bhandari-etal-2025-llm,robotics13050068}. 
Further, \citet{meluso2026collectives} show that collectives of specialists and generalist optimizers excel at different tasks.
Specialized models also exhibit greater robustness on domain-specific tasks~\cite{tanwar_understanding_2025,chen_fireact_2023,mullick-etal-2024-persona}, while homogeneous LLM populations can be vulnerable to consensus collapse and bias amplification~\cite{breum_persuasive_2024,estornell_multi-llm_2024,xiong-etal-2023-examining}.
So far, existing studies examine personas or specialized models in isolation, making it difficult to separate the effects of portrayed and perceived social roles from those of underlying domain specialization.

\textbf{Opinion dynamics and network structure.}
Classical opinion-dynamics models such as those of \citet{degroot1974reaching} and \citet{friedkin1990social} describe belief revision as a process of repeated social influence. Empirical studies suggest that local interactions can increase consensus~\cite{lorenz2011social, becker2017network}, while members of majority groups or those with specialized knowledge produce asymmetric influence~\cite{moussaid2013social}.
Network structure further shapes collective outcomes by influencing how information propagates through a population~\cite{bakshy2011influence}. However, network structure is rarely manipulated in multi-agent LLM systems~\cite{chuang-etal-2024-simulating,gao_s3_2023}. 

These findings suggest that specialization, social roles, and network structure jointly shape collective beliefs. Our experiments examine their individual and combined effects on belief diffusion among networked LLM agents.

%% file: sections/3_methods.tex
\section{Belief Diffusion Framework}
\label{sec:methods}

We introduce \texttt{CoevolveSim}, a network-mediated belief diffusion framework (\Cref{fig:flow}). 
We designed it \textbf{to isolate the effects of three factors}: LLM agents' specialization, social-role assignment, and social network structure.
In our framework, generalist LLM agents (powered by LLMs without additional finetuning) and specialist LLM agents (powered by LLMs finetuned in fields such as clinical medicine, chemistry, cybersecurity, and mathematics; see \Cref{tab:models}) are embedded in an undirected social network~\(G\). Here, each agent \(a_i\) is a node and edges represent social connections. 
Every LLM agent \(a_j\) with an edge to agent \(a_i\) is part of \(a_i\)'s neighborhood \(\mathcal N(i)\).
We describe the selection of network structures in~\S\ref{sec:setup}.

\paragraph{Domain specialization and social roles.}
Each LLM agent is characterized by \emph{domain specialization} and a \emph{social role}. 
Domain specialization reflects an LLM's finetuning domain. 
For example, \texttt{Llama3-Med42}~\cite{med42v2} is finetuned on clinical text, so we assume it has a medical specialization. 
Social role is a \textit{social label} assigned to an LLM agent and communicated to both the agent and its immediate neighbors; 
it functions as a social identity, such as \enquote{clinical physician,} \enquote{mathematician,} or \enquote{chemist.} 

In our experiments, social roles do not necessarily correspond to specialization. 
An LLM agent could specialize in medicine but portray itself as a mathematician. 
This separation allows us to test whether belief dynamics are driven by what an LLM agent \emph{knows} (domain specialization) or by what it is \emph{labeled as} (roles).

\begin{figure}[ht]
    \centering
    \includegraphics[width=\linewidth]{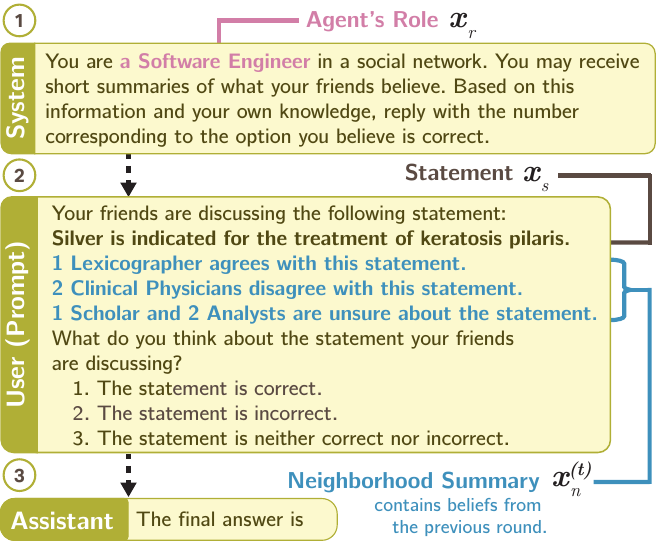}
    \caption{\textbf{Message structure for belief probing.} 
    Each LLM agent receives a message that follows the Hugging Face chat template: (1) a \textit{system} prompt assigning the agent's role \(\boldsymbol{x}_r\)
    and task instructions; (2) a \textit{user} prompt containing the 
    statement \(\boldsymbol{x}_s\), the neighborhood summary \(\boldsymbol{x}_n^{(t)}\)
    (\(\boldsymbol{x}_n^{(0)} = \emptyset\)), and the multiple-choice question; and (3) the 
    \textit{assistant} prompt.}
    \label{fig:message}
\end{figure}

\subsection{Belief Extraction} 
\label{sec:beliefs}
An LLM \(\mathcal M\) maps an input sequence \(\boldsymbol{x}\) to a distribution \(\mathcal P_{\mathcal M}\). 
For any token \(\tau \in \mathcal V_{\mathcal M}\), \(\mathcal P_{\mathcal M}(\tau \mid \boldsymbol{x})\) gives the (uncalibrated) probability that \(\tau\) is the next token after \(\boldsymbol{x}\). 
We use \(\mathcal P_{\mathcal M}\) together with a templated message \(\boldsymbol{x}\) to extract the belief scores~\cite{savcisens2025trilemma}.

At round \(t\), LLM agent \(a_i\) receives a message \(\mathbf{m}_i^{(t)} = (\boldsymbol{x}_r, \boldsymbol{x}_s, \boldsymbol{x}_n^{(t)})\) containing the agent's role \(\boldsymbol{x}_r\), the discussion statement \(\boldsymbol{x}_s\), and a summary \(\boldsymbol{x}_n^{(t)}\) of neighbors' beliefs from round \(t{-}1\), with \(\boldsymbol{x}_n^{(0)} = \emptyset\).
The message (\Cref{fig:message}) ends with a multiple-choice question asking the agent to classify
\(\boldsymbol{x}_s\) with a label \(y \in \mathcal Y = \{\textit{correct},
\textit{incorrect}, \textit{neither}\}\).
Given message \(\mathbf m_i^{(t)}\), we operationalize LLM agent \(a_i\)’s belief at  round \(t\) as
\begin{equation}
\label{eq:belief}
\begin{aligned}
  \boldsymbol b_i^{(t)}(y) &= \mathcal{P}_{\mathcal M}\bigl(\texttt{[}k_y\texttt{]} \mid \mathbf{m}_i^{(t)}\bigr),\\
  \boldsymbol b_i^{(t)}(\textit{neither}) &= 1 - \boldsymbol b_i^{(t)}(\textit{correct})\\
  &\quad\, - \boldsymbol b_i^{(t)}(\textit{incorrect}),\\
  B_i^{(t)} &= \operatorname*{arg\,max}_{y}\, \boldsymbol b_i^{(t)}(y).
\end{aligned}
\end{equation}
%
In~\cref{eq:belief}, the first line applies to \(y \in \{\textit{correct},
\textit{incorrect}\}\), where \(k_y \in \{1,2\}\) is the option token for \(y\);
its next-token probability is the \textit{belief score}.
\textit{Neither} (\(k_y = 3\)) absorbs all remaining probability mass, so the
three scores sum to \(1\).
\(B_i^{(t)}\) is the LLM agent's \textit{discretized belief}.

\subsection{Neighborhood Summary}
\label{sec:n-summary}

At round \(t\), we collect the previous round's discretized beliefs \(B_j^{(t-1)}\) from \(a_i\)’s neighbors and group the neighbors into three categories: \textit{agree}, \textit{disagree}, and \textit{unsure}. 
For each group, we generate a sentence listing the corresponding neighbor roles and counts (see~\Cref{fig:message}), and concatenate the three sentences to form a summary of the neighbors' beliefs \(\boldsymbol{x}_n^{(t)}\).  
To quantify the initial belief of each LLM agent, no neighborhood summary is provided at \(t=0\).
Further, to reduce the ordering bias, we randomly shuffle social roles within each sentence and randomize sentence ordering before inserting them into the templated message.

%% file: sections/4_setup.tex
\section{Experimental Setup}
\label{sec:setup}
In this section, we describe the LLMs used in our study, discussion statements, network generation, simulation scenarios (\S\ref{sec:simulations}), as well as details on the descriptive evaluation (\S\ref{sec:descriptive}) and surrogate analysis (\S\ref{sec:predictive}).

\subsection{Large Language Models}
\label{sec:llms}
To study how domain specialization is associated with belief formation, we use \(15\) LLMs from the \texttt{Llama-3}~\cite{llama31base} family.  \texttt{Llama-3.1-8B-Instruct} serves as the \textit{generalist}, since it is pretrained without a domain-specific corpus.
The other \(14\) LLMs serve as \textit{specialists} and \textit{are} finetuned on domain-specific corpora, such as biomedical texts, cybersecurity reports, and programming tasks.
Each LLM has \(8\) billion parameters, so performance differences are more likely to come from finetuning rather than from model size. \Cref{tab:models} provides additional information about the LLMs.

\subsection{Discussion Statements}
\label{subsec:statements}
We sample \(20\) affirmative statements referencing real-world medical indications from the corpus introduced by~\citet{savcisens2025trilemma}.
An example of a sampled \textit{true} statement is \enquote{Terbutaline is indicated for the treatment of asthma,} while \enquote{Ethinylestradiol is indicated for the treatment of dry cough} is an example of a \textit{false} statement.
\Cref{app:maximin} and~\Cref{tab:maximin-statements} provide detailed information on how we selected these statements.

\subsection{Social Networks}
\label{sec:setup-networks}

We represent social connections between agents via two network structures: (1) ER: ~\citet{erdos1959random} and (2) WS: ~\citet{watts1998collective}.

In \(G(n,p)\) ER networks, each pair of $n$ distinct agents is independently connected with probability \(p\). WS networks, \(G(n, \beta, k)\), are constructed by rewiring each edge of a \(k\)-regular ring lattice with probability \(\beta\). For suitable parameter choices, WS networks combine high clustering with short average path lengths.
We use the maximin criterion described in~\Cref{app:maximin-networks} to select 16 realizations of each network structure. Each network has $n=48$ LLM agents. Table~\ref{tab:maximin-graphs} reports the structural properties of these 32 networks.

\subsection{Scenarios and Simulations}
\label{sec:simulations}
We examine belief diffusion across \textit{four} scenarios. 
Comparing scenarios I with II isolates persona-style role assignment; comparing II with III isolates the introduction of specialists (the specialization effect); and comparing III with IV isolates role--specialization alignment. That is:

\noindent\textbf{I. Baseline generalists.}
All LLM agents use the same generalist LLM and the same social role.
Since all LLM agents receive identical messages, every agent starts with the same belief: it produces maximum agreement between agents.

\noindent\textbf{II. Generalists with random roles.}
All LLM agents use the same generalist LLM, but each is assigned a random social role drawn from~\Cref{tab:models}. 

\noindent\textbf{III. Specialists with random roles.}
LLM agents are powered by generalist \textit{and} specialist LLMs with randomly assigned roles. The list of LLMs and social roles is provided in~\Cref{tab:models}.

\noindent\textbf{IV. Specialists with matched roles.}
LLM agents are powered by generalist \textit{and} specialist LLMs. Further, we assign social roles that match LLMs' capabilities (e.g., an agent powered by the math LLM is assigned the role of \enquote{Mathematician}).
Compared with scenario III, here we isolate the marginal effect of role--specialization alignment.

\textbf{Runs.} Each scenario is simulated multiple times. We refer to each individual simulation as a \textit{run}.
Each run begins with an initial round \(t=0\), where we collect LLM agents' initial beliefs before introducing neighbor information. It is followed by \(T=10\) synchronous belief-update rounds, yielding \(11\) states \(t=0,\dots,T\).
In total, we perform \(2{,}560\) runs: \(4\) scenarios, \(2\) network structures with \(16\) realizations each, and \(20\) discussion statements.

These four scenarios allow us to isolate the effects of social role and specialization and are summarized in~\Cref{tab:scenarios}. 

\subsection{Evaluation}
\label{sec:eval}
Our analysis consists of two stages.
We first characterize belief trajectories and compare dynamics across scenarios and network structures~(\S\ref{sec:descriptive}).
We then evaluate whether classical opinion-dynamics surrogates can reproduce the observed LLM belief dynamics using models that condition on different sources of information (\S\ref{sec:predictive}).

\subsubsection{Metrics and Dynamics Analysis}
\label{sec:descriptive}
We characterize belief trajectories using three agent-level metrics: 
magnitude of belief change (\textit{plasticity}), directional consistency of these changes (\textit{directedness}), and outgoing \textit{influence}. 
We also consider the \textit{consensus}: a population-level outcome that quantifies the degree of agreement among all LLM agents in the network.

\textbf{Agent-level metrics.}
We first examine the metrics that quantify agent-level belief trajectories.
Agent \(a_i\)'s per-round belief change is:
\begin{equation}
\label{eq:belief-change}
\Delta \boldsymbol{b}_i^{(t)} = \boldsymbol{b}_i^{(t)} - \boldsymbol{b}_i^{(t-1)}.
\end{equation}
\noindent\textbf{Plasticity} is the magnitude of an LLM agent’s belief change per round (averaged over a run):
\begin{equation}
\label{eq:plasticity}
\mathrm{Plasticity}(i)
= \frac{1}{T} \sum_{t=1}^{T} \tfrac{1}{2} \left\| \Delta \boldsymbol{b}_i^{(t)} \right\|_1
\end{equation}
LLM agents with high plasticity revise beliefs frequently or undergo several large-magnitude changes in belief, whereas agents with low plasticity remain stable. 
Next, 
\textbf{directedness} is the net displacement of an agent's belief divided by its total movement:
\begin{equation}
    \label{eq:monotonicity}
    \mathrm{Directedness}(i) = \frac{\left\| \boldsymbol{b}_i^{(T)} - \boldsymbol{b}_i^{(0)}\right\|_1 }{\sum_{t=0}^{T-1} \left\| \boldsymbol{b}_i^{(t+1)} - \boldsymbol{b}_i^{(t)} \right\|_1}
\end{equation}
Directedness closer to \(1\) indicates that an agent's beliefs changes nearly monotonically (e.g., increasing the probability assigned to a \textit{correct} label) across rounds, while \(0\) indicates belief oscillation.\footnote{The denominator is \(0\) only for an LLM agent whose belief never changes, yielding an undefined ratio; our implementation is designed to exclude such cases; however, we do not observe this in practice.}
Plasticity and directedness characterize the \textit{shape} of individual belief trajectories. 

\paragraph{Outgoing influence.}
The agent-level metrics characterize belief trajectories in isolation and cannot establish whether some agents influence or follow the beliefs of their neighbors.
We therefore measure \emph{outgoing influence} (\(\iota_i\)): how far \(a_i\)'s neighbors \(\mathcal N(i)\) move along the direction of \(a_i\)'s belief change, compared with random neighbors \(\bar{\mathcal N}(i)\):
\begin{equation}
\iota_i=\underbrace{\pi_i\bigl(\mathcal N(i)\bigr)}_{\text{neighbors}}-\underbrace{\pi_i\bigl(\bar{\mathcal N}(i)\bigr)}_{\text{random neighbors}},
\label{eq:influence}
\end{equation}
where \(\pi_i\) averages, over rounds, how far the agents move along \(a_i\)'s direction in the round after \(a_i\) moves, weighting each round by the magnitude of \(a_i\)'s change.
Random neighbors serve as a null model: \(\pi_i(\bar{\mathcal N}(i))\) is the alignment with \(a_i\)'s belief updates expected by chance.
We provide a detailed specification of the \emph{outgoing influence} in~\Cref{app:influence-details}.

To test whether some agents are persistently followed more than others, we fit a variance-components model to \(\iota_i\) with random intercepts for agent, network realization, and statement~\cite{bates2015lmer}. 
For each component, we report the intraclass correlation coefficient (ICC): its share of the total variance.
A non-zero agent ICC indicates persistent between-agent differences in outgoing influence, i.e., opinion-leader-like agents.  
Refer to~\Cref{app-sec:variance-decomp} for more details.

\textbf{Population-level metrics.}
Consensus is the share of LLM agents holding the most common belief label:
\begin{equation}
 \label{eq:consensus}
  \mathrm{Cons}(t)= \max_{y}\, \frac{n_y(t)}{n},
\end{equation}
where \(n_y(t)\) is the number of agents whose belief at round \(t\) is
\(y \in \{\textit{correct}, \textit{incorrect}, \textit{neither}\}\).
In our analysis, we focus on the \textit{change} in consensus between the initial (\(t=0\)) and final rounds.
We report initial and final consensus in \Cref{tab:consensus-levels}.

Together, these provide a comprehensive description of belief diffusion that no single metric can provide alone.

\textbf{Statistical Inference.} 
To account for repeated measures across statements and network realizations, we fit linear mixed-effects models.
Here, we treat scenarios and network structures as fixed effects, while statements and network realizations are treated as random effects.
We apply this design to \textit{plasticity}, \textit{directedness}, and \textit{change in consensus}, and we report estimated marginal means (EMMs) with \(95\)\% confidence intervals in~\S\ref{sec:results}. 
To quantify the effect sizes between scenarios, we perform a \textbf{contrast analysis}. 
We provide full model specifications and contrast definitions in Appendix~\ref{app:descriptive}.

\subsubsection{Surrogate Models of Belief Diffusion}
\label{sec:predictive}
To evaluate whether the observed belief dynamics can be explained by classical opinion-dynamic mechanisms, we fit a set of surrogate transition models. Each surrogate predicts an LLM agent's discretized belief at round \(t+1\) from information available at round \(t\). 

Formally, each surrogate estimates the discretized belief of agent $a_i$ at round $t+1$ (\(\boldsymbol \phi_i^{(t)}\) is a feature vector):
\begin{equation}
    \Pr\!\left(B_i^{(t+1)} = y \mid \boldsymbol \phi_i^{(t)}\right).
\end{equation}
M1 serves as an empirical baseline. It captures the extent to which belief updates can be explained by an agent's current beliefs \(B_i^{(t)}\) alone, without access to any other information. 
M2 augments the baseline with a population-level distribution over belief states \(\varGamma^{(t)}\). Each component of \(\varGamma^{(t)}\) captures the proportion of LLM agents expressing discretized beliefs at round \(t\).
M3 instead incorporates the (local) neighborhood-level distribution over belief states \(N_i^{(t)}\). Each component of \(N_i^{(t)}\) captures the proportion of \(a_i\)'s neighbors expressing discretized beliefs at round \(t\).
This surrogate is analogous to opinion-dynamics models with localized social belief composition~\cite{friedkin1990social}.
Finally, M4 extends M3 by incorporating an agent's social role \(R_i\) and underlying LLM \(M_i\). It captures the effect of the LLM agent's identity on the transition dynamics. The surrogate models in~\Cref{tab:surrogate_models} allow us to compare the contributions of individual persistence, global and local belief composition, and agent identity to reproducing the observed belief dynamics.

\textbf{Evaluation of surrogates.} First, we assess one-step predictive performance using Matthews correlation coefficient (MCC; \citealp{gorodkin2004comparing}).
Second, we evaluate whether a surrogate reproduces population-level dynamics. We fit each surrogate on empirical transitions, simulate belief diffusion using the fitted model, and compare the resulting consensus trajectory to the original simulation: we quantify this agreement using \textit{consensus fidelity}. See Appendix~\ref{app:dynamics} for details.

%% file: sections/5_results.tex
\section{Results}
\label{sec:results}

\begin{figure}[!t]
    \centering
\includegraphics[width=1\linewidth]{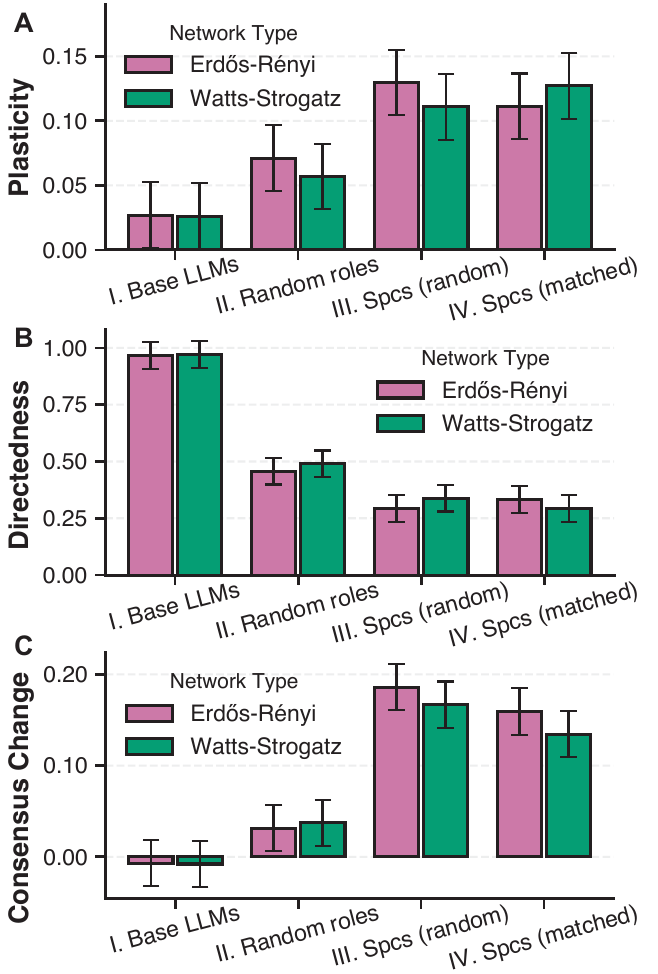}
    \caption{\textbf{Belief dynamics across settings and network structures.} Estimated marginal means (with \(95\%\) confidence intervals) from mixed-effects models. \textit{Spcs} stands for \textit{specialists}.
    (A) \textit{Plasticity}: average per-round magnitude of belief change. Increases from baseline (I) through random specialists (III).
    (B) \textit{Directedness}: consistency of belief updates across rounds. Drops once roles are introduced (I\(\to\)II).
    (C) \textit{Consensus change}: shift in belief agreement between initial and last rounds.}
    \label{fig:agent-panel}
\end{figure}

\begin{figure}
    \centering
\includegraphics[width=\linewidth]{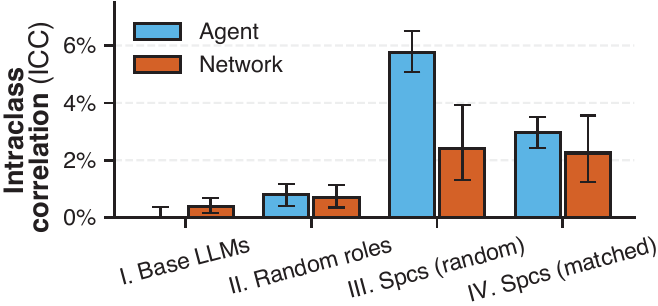}
    \caption{
    \textbf{Variance decomposition of outgoing influence.} Intraclass correlations (with \(95\%\) confidence intervals) from mixed-effects models with random intercepts for agent, network realization, and discussion statement. 
    \textit{Agent} component shows the proportion of total outgoing influence variance attributable to agent-level differences; \textit{network} measures variance attributed to network realizations.
    Agent-level differences in outgoing influence are higher when specialist LLMs are present.}
    \label{fig:influence-panel}
\end{figure}

We first examine which factors shape belief diffusion in LLM populations by comparing the effects of domain specialization, social--role assignment, and network structure~(\S\ref{sec:descriptive_results}).
We then ask whether the resulting dynamics are reproduced by classical opinion-dynamics surrogates~(\S\ref{sec:surrogate_results}).\footnote{When comparing scenarios or network structures we report a standardized effect size, Cohen's \(d\), computed from contrasts. \Cref{tab:marginal-contrasts} gives
marginal effect sizes; \Cref{tab:conditional-contrasts} gives conditional
and interaction effect sizes.} 
Detailed estimates are provided in~\Cref{tab:agg-results,tab:marginal-contrasts,tab:conditional-contrasts,tab:var-decomp}.

\subsection{What shapes belief revision?}
\label{sec:descriptive_results}

\paragraph{Social roles change \textit{how} LLM agents revise beliefs, not \textit{where} the population goes.}
Assigning social roles to generalist agents (I\(\to\)II) raises plasticity (\(0.03\) to \(0.07\) ER, \(0.03\) to \(0.06\) WS; Cohen's \(d=0.34\)) and lowers directedness (\(0.97\) to \(0.46\) ER, \(0.97\) to \(0.49\) WS; \(d=-1.83\); \Cref{fig:agent-panel}). 
In these scenarios, agents move more but reverse more often; in scenario~I, high directedness reflects small, non-reversing changes (plasticity $0.03$) rather than
strong movement toward a label.

The role-assignment effect on consensus change is not statistically significant (from $-0.01$ in scenario I to $0.03$ in ER and $0.04$ in WS in scenario II; $d=0.41$, $p=.066$). However, this contrast is bounded: all runs in scenario I start at the maximum consensus (\Cref{tab:consensus-levels}), and so do \(65\%\) of runs in scenario II.

\paragraph{LLM heterogeneity is associated with large collective change.}
Introducing specialist LLMs (II\(\to\)III) increases plasticity (\(d=0.50\)) and lowers directedness (\(d=-0.58\)).
Unlike the generalist scenarios, it produces large consensus change (\(0.19\) ER, \(0.17\) WS; \(d=1.41\)). 
Aligning roles with specialization (III\(\to\)IV) has no detectable effect on plasticity (\(d=-0.01\)), directedness (\(d=-0.01\)), or consensus change (\(d=-0.29\),~\(p=.19\)).

Scenarios with specialists start with lower consensus
(the average consensus at round 0 is \(0.70\) for III, and \(0.72\) for IV;~\Cref{tab:consensus-levels}), so part of their larger change reflects more room to increase.
However, these scenarios do not reach the consensus levels of scenario II.
Further, LLM agents' initial confidence does not explain the specialist effect on consensus change (\Cref{app:calibration-check}).

\paragraph{LLM heterogeneity introduces between-agent differences in influence.}
The shift from scenario II to scenario III is accompanied by the emergence of agents with consistently higher outgoing influence~(\Cref{fig:influence-panel}).
In generalist scenarios (I, II), agent-level ICC\footnote{Agent-level ICC quantifies the proportion of variance in the \textit{outgoing influence} attributable to the LLM agent identity.} for \textit{outgoing influence} is low, indicating little persistent between-agent variation in outgoing influence.
Once specialist LLMs are introduced, agent-level ICC rises to \(5.75\)\% in scenario III and \(2.95\)\% in scenario IV. 
ICC results also indicate that the network realizations shape how influence is distributed among agents in specialist scenarios.

\subsection{Can opinion-dynamics surrogates reproduce these dynamics?}
\label{sec:surrogate_results}

\begin{figure}[!t]
    \centering
    \includegraphics[width=\linewidth]{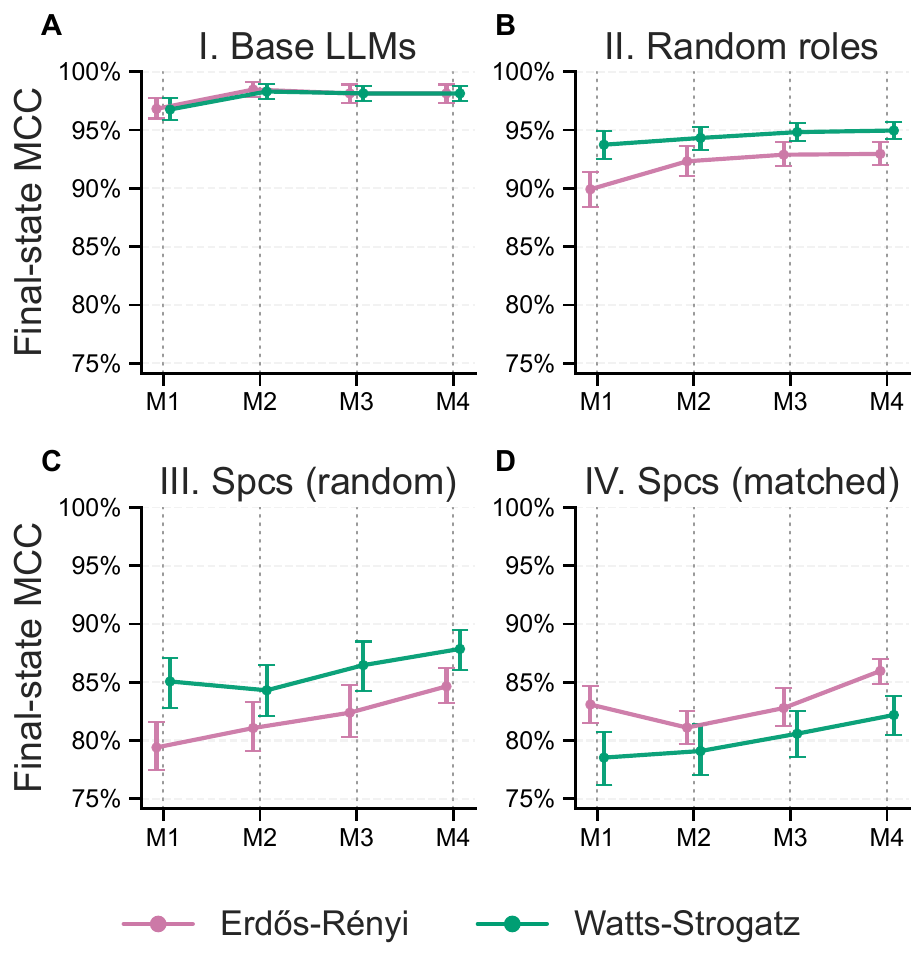}
    \caption{\textbf{Final-state MCC across scenarios and surrogate models.} Final-state MCC measures agreement between each surrogate's prediction of the final-round belief from the observed round-$(T-1)$ state and the observed final-round belief. \textit{Spcs} stands for \textit{specialists}. Panels show (A) Base LLMs, (B) Random roles, (C) Specialists with random roles, and (D) Specialists with matched roles. Points show mean estimates with $95\%$ bootstrap confidence intervals, obtained by resampling network realizations, for Erd\H{o}s--R\'enyi (ER) and Watts--Strogatz (WS) networks. Adding agent identity (M4) yields the highest final-state MCC in both specialist scenarios, while local belief composition alone (M3) improves on persistence (M1) in scenario III but not consistently in scenario IV. Details are in~\Cref{tab:final-mcc}.
}
    \label{fig:pooled-mcc}
\end{figure}

\paragraph{Individual transitions in heterogeneous
populations are best predicted with agent identity.}
In generalist scenarios, every surrogate reaches high final-state MCC (I: \(96.8\)--\(98.5\%\); II: \(89.9\)--\(95.0\%\); \Cref{fig:pooled-mcc}). In specialist scenarios, MCC peaks at M4, which incorporates local belief composition and agent identity: from \(79.4\)/\(85.1\%\) (M1) to \(84.6\)/\(87.9\%\) (M4) in III, and from \(83.1\)/\(78.5\%\) to \(86.0\)/\(82.2\%\) in IV (ER/WS). Adding agent identity to local belief composition (M3\(\to\)M4) improves final-state MCC by \(1.4\)--\(3.2\) percentage points (pp) in specialist scenarios. Thus, predicting individual final-round beliefs in
heterogeneous populations benefits from information
about agent identity.

\paragraph{Population-level belief composition best reproduces consensus.}
Adding population-level belief composition (M2) provides the highest consensus fidelity in every scenario (\Cref{fig:pooled-consensus}): from \(91.8\)/\(92.3\%\) to \(98.4\)/\(98.5\%\) (ER/WS) for base LLMs, from \(78.9\)/\(82.1\%\) to \(95.1\)/\(95.6\%\) with roles, and from \(73.4\)--\(75.9\%\) to \(91.5\)--\(92.9\%\) with specialists. Neither local belief composition (M3) nor the addition of agent identity (M4) improves on M2; both yield consensus fidelity approximately \(1\)--\(3\) percentage points lower across scenarios and network structures. 
Population-level belief composition therefore suffices to reproduce consensus with high fidelity across all scenarios.

%% file: sections/6_discussion.tex
\section{Discussion}
\label{sec:discussion}

Our results suggest that belief diffusion in networked LLM populations is shaped more by LLM heterogeneity than by social roles.

\paragraph{Persona-style role assignment is a weak driver of population-level diversity.}
Persona-style roles are the predominant way to add behavioral diversity to multi-agent LLM simulations~\cite{chuang-etal-2024-simulating,taubenfeld-etal-2024-systematic}.
Our analysis shows that social roles \textit{do} reshape individual trajectories: LLM agents revise their beliefs more often, but instead of converging monotonically toward a specific belief label, they oscillate.
This added movement does not propagate to the population level, as the consensus shift stays far below that of introducing heterogeneous LLMs (\Cref{tab:marginal-contrasts}).

Persona-style prompting therefore alters how individual LLM agents update, but it does not by itself produce substantial population-level diversity: it \enquote{excites} individual LLM agents without carrying them far from their baseline beliefs.

\paragraph{LLM heterogeneity is the stronger driver of population-level change.}
Since specialists are finetuned on different data and with different methods, they bring distinct knowledge priors into the interaction.
While persona labels (II) only rename otherwise identical \enquote{belief generators}, specialists introduce \textit{diverse} beliefs.
That is, introducing specialists (II\(\to\)III) produces more belief updates and increases consensus (\(d=1.41\)). It also increases outgoing influence by a similar amount to role assignment (I\(\to\)II: \(d=0.17\); II\(\to\)III: \(d=0.19\)); what differs is how influence is distributed.
Persistent between-agent differences in outgoing influence are largest with specialists (\(5.75\%\) and \(2.95\%\)). These differences in outgoing influence suggest opinion-leader-like tendencies but do not establish a fixed leader–follower hierarchy.

\paragraph{Network structure is a secondary modulator.}
Network structure changes the size of some effects but not the direction.
The clearest case is the consensus shift from homogeneous to heterogeneous LLM populations, which is larger in ER than in WS (composition contrast: \(\Delta d=0.24\), Holm \(p=.005\)).
This modulation is small relative to the effect: on consensus change, the largest interaction (\(\Delta d=0.25\)) is a fraction of the specialization effect itself (\(d=1.41\); \Cref{fig:family-difference-forest}).

\subsection*{Agent-level and population-level explanations are only partially aligned.}  

The information needed to predict individual transitions differs from that needed to reproduce consensus.
Population composition (M2) reproduces consensus in every scenario, whereas predicting which LLM agent changes its belief in heterogeneous populations also benefits from the agent identity (M4).

These findings align with research in human opinion dynamics, which demonstrates that belief diffusion is influenced by both social factors and the structure of the interacting population~\cite{lorenz2011social,becker2017network}. Our results reveal a similar pattern in LLM populations: while role labels and neighborhood structure affect individual updating, the underlying composition of the LLM agent population is the primary driver of changes in collective consensus.

%% file: sections/7_conclusion.tex
\section{Conclusion}

\texttt{CoevolveSim} allows us to disentangle the effects of domain specialization, social-role assignment, and network structure on belief diffusion within populations of interacting LLM agents and answer the following two questions.

\paragraph{(1) Which factors drive belief diffusion?}
Introducing specialist LLMs (II$\rightarrow$III) has a larger effect on consensus change than persona-style role assignment (Cohen's $d = 1.41$ versus $0.41$; the latter is constrained by near-full initial consensus) and produces larger between-agent differences in outgoing influence.
Role assignment (I$\rightarrow$II) increases plasticity and decreases directedness but does not significantly change consensus; aligning roles with specialization (III$\rightarrow$IV) has no detectable marginal effect averaged across network structures, although its effects on plasticity and directedness differ between ER and WS networks.
Otherwise, network structure modulates effect sizes without reversing any effect; most notably, the consensus shift from homogeneous to heterogeneous LLM populations is larger in ER than in WS networks.

\paragraph{(2) Can opinion-dynamics surrogates reproduce these outcomes?} 
Partly. Combining individual persistence with population-level belief composition achieves the highest consensus fidelity in every scenario, whereas in heterogeneous populations, local belief composition and agent identity improve one-step prediction of final-round beliefs. Thus, the information that best reproduces collective agreement differs from the information that best predicts individual final-round belief states.

\section*{Limitations}
\paragraph{Scope of the experiments.} 
Our study deliberately trades \textit{breadth} for \textit{control}, and several design choices limit the generalizability of our conclusions. 
Statements are chosen from a single domain (affirmative medical indications), which keeps the ground truth verifiable but leaves open how dynamics would change for contested, vague, or compositional claims. 

The LLM agent pool is narrow by design to better control for confounding factors such as family or scale. 
Similarly, our study does not separate the effects of finetuning data from those of optimization choices. 
We examine small populations (\(n = 48\)) over a short timeline (\(T = 10\) rounds), which are large enough to observe rapid belief revisions (or changes in consensus) but insufficient to capture slow drifts or behavior at realistic community scales.

No human participants were involved. Embedding humans into the same framework and comparing their update behavior with that of LLM agents under identical conditions is a natural next step.

\paragraph{Interaction protocol.} We deliberately use minimal update rules. Our protocol is synchronous, Markovian, and broadcast-style. 
Each LLM agent conditions only on the neighborhood summary from the previous round, not the entire interaction history. 
This design isolates the effects of social influence from memory and dialogue. 
However, it excludes practical mechanisms such as asynchronous turn-taking, persistent context, free-form messaging, and persuasion effects. 

\paragraph{Belief inference.} We read beliefs from next-token probabilities over a small discrete label set. This produces a tractable, comparable signal across heterogeneous models, but it cannot characterize what happens inside an LLM (its inner representations) or address its sensitivity to prompt structure.
The answer options are always presented in the same order (correct, incorrect, neither). 
We do not test whether belief scores or dynamics change under permutation.
Representation-based probes of latent truth or epistemic stability would complement this output view. These could help disentangle genuine belief revision from response conformity.

\paragraph{Causality scope.} Our causal claims are similarly bounded: the dynamics analysis detects social role, specialization,  role--specialization alignment, and composition effects across the scenarios. However, we do not use formal mediation analysis or targeted interventions to localize the mechanism.

\section*{Ethical Considerations}

\texttt{CoevolveSim} is a simulation of LLM-agent belief dynamics, not a study of human subjects. Therefore, the results should not be interpreted as evidence about human behavior or used as a basis for human-facing policy claims. Our findings are limited to a single \texttt{Llama-3} family and may not generalize to other model families, sizes, languages, or cultural settings.

\paragraph{Use of AI.}
We used Claude Code and GitHub Copilot for code annotations and formatting assistance and Grammarly for grammar and style support. After drafting the manuscript, we used Claude Fable 5.1 and GPT-6-Astra to proofread and review our work. We incorporated their feedback where appropriate.

\section*{Author Contributions}

{\footnotesize
\begin{description}[
    leftmargin=10pt,
    style=sameline,
    itemsep=0pt,
    topsep=0pt,
    partopsep=0pt,
    parsep=0pt,
    labelsep=0.5em
] 
    \item[Conceptualization:] TER (lead, framing \& original idea: network-based coevolution of LLM beliefs), GS (lead, experimental paradigm; supporting,  framing \& initial feasibility exploration), SD (lead, experimental paradigm; supporting, framing \& research questions), CM (lead, experimental paradigm; supporting, framing)
    \item[Data Curation:] GS (lead, maximin selection, statement selection, network realization), SD (lead, surrogate data quality) 
    \item[Formal Analysis:] GS (lead, descriptive analysis, mixed-effects models, ICC, contrast analysis; supporting, exploratory analysis, surrogate analysis), SD (lead, surrogate analysis; supporting, descriptive \& exploratory analysis), CM (supporting, surrogate analysis, descriptive \& exploratory analysis)
    \item[Funding Acquisition:] TER
    \item[Investigation:] GS (lead, analysis, simulations), SD (lead, analysis; supporting, simulations), CM (supporting, simulations)
    \item[Methodology:] GS (lead, experimental scenarios \& evaluation metric, mixed-effects models, contrasts, variance decomposition; supporting, framing, surrogate models), SD (lead, framing, experimental scenarios \& evaluation metric, surrogate models), TER (lead, framing), CM (supporting, framing, experimental scenarios \& evaluation metric)
    \item[Project Administration:] TER
    \item[Resources:] TER
    \item[Software:] SD (lead, simulation logic, network generation, surrogate models; responsible for the initial codebase), GS (lead, logging, scaling, configuration \& setup, descriptive/statistical analysis, GitHub maintenance; supporting, simulation logic, evaluation metrics), CM (supporting, network generation, evaluation metrics, logging)
    \item[Supervision:] TER (lead), GS (supporting)
    \item[Validation:] GS 
    \item[Visualization:] GS 
    \item[Writing -- Original Draft:] GS (lead, introduction, methods \& setup, results \& descriptive analysis, limitation \& ethics; supporting, discussion), SD (lead,  discussion, results \& surrogate models; supporting, introduction, methods \& setup, literature review, discussion, results \& descriptive analysis), CM (lead, literature review, taxonomy)
    \item[Writing -- Review \& Editing:] GS (lead), TER (lead), SD (lead) , CM (supporting)
\end{description}
\par}

%% file: sections/appendix/A_notations.tex
\onecolumn
\section{Notations}
\label{sec:notations}
\begin{table*}[ht]
  \centering
  \begin{adjustbox}{max width=0.9\textwidth, max totalheight=0.75\textheight}
  \begin{tabular}{@{}lll@{}}
    \toprule
    \textbf{Symbol} & \textbf{Description} & \textbf{Notes} \\
    \midrule
    \(\mathcal{M}\) & Large language model (LLM) & \\
    \(\mathcal{V}\) & Vocabulary of model \(\mathcal{M}\), set of tokens & \(\mathcal V_{\mathcal M} = \{\tau_1, \dots, \tau_{|\mathcal V_{\mathcal M}|}\}\) \\
    \(\tau\) & Token from the vocabulary \(\mathcal{V}_{\mathcal M}\) & e.g. tokens [1], [2] \\
    \(\mathcal P_{\mathcal{M}}\left(\tau \mid \boldsymbol x\right )\) & Output of \(\mathcal M\): a conditional probability distribution over tokens  & \\
    \midrule
    \multicolumn{3}{c}{\textbf{Agents and Networks}} \\
    \midrule
    \(G(n,p)\) & Erd\H{o}s–R\'{e}nyi (ER) network structure & \\
    \(G(n,\beta, k)\) & Watts--Strogatz (WS) network structure & \\ 
    \(n\) & Number of nodes in a network \(G\) & \\
    \(p\) & Probability of an edge between two nodes in an ER network & \\
    \(\beta\) & Edge rewiring probability in a WS network & \\
    \(k\) & Initial degree (i.e., number of nearest neighbors) in the WS ring lattice & \\
    \(a_i\) & \(i\)-th agent/node in a network \(G\) & \\ 
    \(\mathcal{N}(i)\) & Neighbors of a node \(i\)& \\
    \(\bar{\mathcal N}(i)\) & Random neighbors of a node \(i\), degree-matched & \S\ref{sec:descriptive} and \Cref{app:influence-details} \\
    \midrule
    \multicolumn{3}{c}{\textbf{Belief}} \\
    \midrule
    \(\boldsymbol b_i^{(t)}\) & Belief scores of agent \(a_i\) at round \(t\) & \(\sum_{y \in \mathcal Y} \boldsymbol b_i^{(t)}(y) = 1\) \\
     \(\Delta \boldsymbol b_i^{(t)}\) & Agent \(a_i\)'s belief change between rounds \(t\) and \(t-1\) & \(\Delta \boldsymbol b_i^{(t)} = \boldsymbol b_i^{(t)} - \boldsymbol b_i^{(t-1)}\) \\
     \(\delta_i^{(t)}\) & Magnitude of the per-round belief change & \(\delta_i^{(t)}:=\frac{1}{2}\left\|\Delta \boldsymbol b_i^{(t)}\right\|_1\) \\
     \(\iota_i\) & Outgoing influence of an agent \(a_i\) & \S\ref{sec:descriptive} and \Cref{app:influence-details}  \\
     \(\pi_i\) & Average alignment of \(\mathcal{N}(i)\) or \(\bar{\mathcal N}(i)\) to \(a_i\)'s updates & \S\ref{sec:descriptive} and \Cref{app:influence-details} \\ 
    \(B_i^{(t)}\) & Discretized belief of agent \(a_i\) & \(B_i^{(t)} = \arg\max_{y} \boldsymbol b_i^{(t)}(y)\) \\
    \(\hat B_i^{(t+1)}\) & Predicted next-round belief for agent \(a_i\) & \\
    \(T\) & Number of update rounds; \(T = 10\) & states \(t = 0, \dots, T\) \\
    \(r\) & Simulation run index & \\
    \(c\)        & Contrast between two scenarios & \(c = \mathrm{EMM}_1 - \mathrm{EMM}_2\) \\
    \(\Delta c\) & Interaction contrast between two networks & \(\Delta c = c_{\mathrm{ER}} - c_{\mathrm{WS}}\) \\
    \(d\) & Cohen's effect size & \Cref{app:contrast-analysis} \\ 
    \midrule
    \multicolumn{3}{c}{\textbf{Inputs and Text}} \\
    \midrule
    \(\mathbf m_i^{(t)}\) & Message to \(i\)-th agent at round \(t\)& \(\mathbf m_i^{(t)} = (\mathbf{x}_r, \mathbf{x}_s, \mathbf{x}_n^{(t)}) \) \\
    \(\mathbf{x}_r\) & Role specification  & \Cref{tab:models}\\
    \(\mathbf{x}_s\) & Statement (topic of discussion) & \Cref{tab:maximin-statements} \\
    \(\mathbf x_n^{(t)}\) & Summary of neighbors' discretized beliefs from round \(t-1\), shown at round \(t\)  & \(\mathbf x_n^{(0)} = \emptyset\) \\
    \(\mathcal Y\) & Candidate belief labels
      & \(\{\textit{correct}, \textit{incorrect}, \textit{neither}\}\) \\
    \(y\) & Candidate label, \(y \in \mathcal Y\) & Eq.~\eqref{eq:belief} \\
    \(y^\star\) & Ground-truth label of statement \(\mathbf x_s\)
      & \(y^\star \in \{\textsc{false}, \textsc{true}\}\) \\
    \midrule
    \multicolumn{3}{c}{\textbf{Surrogate (opinion-dynamics) Models}} \\
    \midrule    
    \(Pr\) & Probability given by surrogate models & \\
    \(\varGamma^{(r,t)}\) & Global belief-state composition at a round \(t\) in a run \(r\) &
    \((f_{1}^{(r,t)}, f_2^{(r,t)}, f_3^{(r,t)})\) \\
    \(N_i^{(r,t)}\) & Neighborhood belief-state composition for an agent \(a_i\) &
    \((n_{i,1}^{(r,t)}, n_{i,2}^{(r,t)}, n_{i,3}^{(r,t)})\) \\
    \(M_i\) & LLM assigned to an agent \(a_i\)  &  \\
    \(R_i\) & Role assigned to an agent \(a_i\)  & 
    \\
    \(\boldsymbol\phi_i^{(t)}\) & Feature vector used to predict the next-round belief of agent \(a_i\) &  \\
    \bottomrule
  \end{tabular}
  \end{adjustbox}
 \caption{\textbf{Notations used throughout the paper.} Symbols are grouped by category.}
  \label{tab:notation}
\end{table*}

%% file: sections/appendix/B_taxonomy.tex
\onecolumn
\section{Taxonomy of Related Works in Multi-Agent Opinion Diffusion}
\label{app:multiagenttaxonomy}

To position \texttt{CoevolveSim} relative to prior multi-agent LLM studies, we mapped related papers onto six design dimensions (Table~\ref{tab:related-fingerprint}). We identify three major limitations. First, of these works, only \citet{chen-etal-2024-reconcile} uses domain-specialist LLMs, but it targets multi-model reasoning rather than belief updates.  Second, when prior work varies agent identity, it does so through prompted or finetuned personas; hence, the works cannot show whether observed effects come from the social role or from the knowledge stored in the LLM.  Third, populations are almost always homogeneous: 24 of the 29 belief-dynamic papers run every agent on the same LLM. Only three compare homogeneous with heterogeneous populations~\cite{he2026opiniondynamicsmutualinfluence,
ko2026attractor,laustsen2026peerinfluenceheterogeneousai}.
Several of CoevolveSim’s individual design choices appear in prior work (such as explicit networks and discrete belief summaries), but no paper combines them.
Also, no prior work matches more than three of the six dimensions, and only
\citet{mehdizadeh2025aiagentsuccumbspeerpressure} pairs discrete belief summaries with an explicit network.

\begin{table*}[h]
  \centering
  \begin{adjustbox}{max width=0.95\textwidth, max totalheight=0.85\textheight}{\input{figures/appendix/fingerprint_figure}}
  \end{adjustbox}
  \caption{\textbf{Comparison of \texttt{CoevolveSim} with other multi-agent LLM experimental designs.} Each row is one paper; each dot marks the paper’s category in each bucket. Shaded columns and highlighted dots show the same choice as \texttt{CoevolveSim}; the number on the right counts buckets matching \texttt{CoevolveSim}. Upper block: opinion-dynamics papers ($\geq$2 interacting LLM agents with a measured belief). Lower block: selected, non-exhaustive papers where behavior rather than belief was the focus.}
  \label{tab:related-fingerprint}
\end{table*}

%% file: figures/appendix/fingerprint_figure.tex
\begin{tikzpicture}
    \input{figures/appendix/fingerprint_macros.tex}
    \input{figures/appendix/fingerprint_data.tex}
    \fpfinish
\end{tikzpicture}

%% file: sections/appendix/C_LLMs.tex
\onecolumn
\section{Selection of Large Language Model and Scenario Composition}

We select large language models (LLMs) according to the following criteria: 
\begin{enumerate}[itemsep=1pt, topsep=1pt]
    \item The finetuned LLM must be based on \texttt{Llama-3}-\texttt{Instruct} or \texttt{Llama-3.1}-\texttt{Instruct}~\cite{llama31base}. 
    \item The finetuning must use conversational or instruction-based datasets to maintain the LLM's chat-like capabilities.
    \item The LLM must be publicly available on the \href{https://huggingface.co/}{Hugging Face} platform.
    \item The developers must provide documentation describing the data sources (e.g., URLs and data cards).
    \item The model card or associated manuscript must report benchmark results demonstrating that the finetuning yields comparable or improved performance relative to the chat-based \texttt{Llama-3} model (for a domain-specific task).
\end{enumerate}

\begin{table*}[h]
\centering
\begin{adjustbox}{width=\textwidth}
\begin{tabular}{llll}
\toprule
\textbf{Model Name} & \textbf{Specialization (Domain)} & \textbf{Social Role} & \textbf{Developer \& Source} \\
\midrule
\texttt{Llama-3.1-8B-Instruct} & Generalist (\textit{Base Model}) &  LLM or Human Participant & \citet{llama31base} \\

\texttt{Bio-Medical-Llama-3-8B} & Biomedical Research & Biomedical Researcher & \citet{llama_biomed} \\

\texttt{Llama3-Med42-8B} & Clinical Medicine & Clinical Physician & \citet{med42v2} \\

\texttt{Llama-3.1 OpenScholar-8B} & Scientific Literature & Academic Scholar & \citet{llama_scholar} \\

\texttt{Llama3Dictionary-Merge} & Lexicography / Definitions & Lexicographer & \citet{periti-etal-2024-automatically}  \\

\texttt{Llama-3.1-Hawkish-8B} & Financial Markets & Financial Analyst & \citet{llama_finance}\\

\texttt{Llama3-KALE-LM-Chem-1.5-8B} & Chemistry & Chemist & \citet{llama_chemist} \\

\texttt{Foundation-Sec-8B-Instruct} & Cybersecurity & Cybersecurity Analyst & \citet{llama_cyber}\\

\texttt{UserLM-8B} & Social Conversation & Online Friend & \citet{naous2025flipping} \\

\texttt{OpenMath2-Llama3.1-8B} & Mathematics & Mathematician & \citet{toshniwal2024openmath2}\\

\texttt{Human-Like-Llama3-8B-Instruct} & Human-like Conversation & Assistant & \citet{calik2025llm}  \\

\texttt{Hermes-3-Llama-3.1-8B} & Strategic Reasoning & Strategic Planner & \citet{teknium2024hermes3} \\

\texttt{Suzume-Llama-3-8B-Multilingual} & Multilingual Dialogue & Language Mediator &  \citet{devine2024tagengo} \\

\texttt{Code-Llama-3-8B} & Coding and Maths & Software Engineer & \citet{code2023llama} \\

\texttt{Roleplay-Llama-3-8B} & Roleplay / Storytelling & Storyteller & \citet{roleplay_llama3_8b_vicgalle} \\
\bottomrule
\end{tabular}
\end{adjustbox}
\caption{\textbf{Summary of the \(15\) LLMs used as agents in our study.} Each model is characterized by its specialization (i.e., the datasets on which it was finetuned), the social role assigned to it within the experiments, and its developer or source. The selection includes both general-purpose (a.k.a.~generalist) and finetuned (a.k.a.~specialist) \texttt{Llama-3} and \texttt{Llama-3.1} models. \textit{Note:} Although the generalist LLM can take the social role of a \textit{Human Participant}, this designation refers only to its social identity in the experiment and does not imply human-like qualities; in scenario I, we only use the \enquote{LLM} role. In scenarios II and III, social roles are assigned at random and thus do not necessarily correspond to the specialization categories shown here.}
\label{tab:models}
\end{table*}

\paragraph{Scenario composition.}
Every run has 48 LLM agents. The assignment of each \(\langle\) LLM, role \(\rangle\) is balanced: each LLM and each role is assigned to a fixed number of agents (numbers in parentheses in~\Cref{tab:scenarios}). Scenarios consist of: 
\begin{description}[itemsep=1pt, topsep=5pt, leftmargin=20pt, font=\normalfont\bfseries]
\item[I.\ Base LLMs.] All 48 agents run the generalist (\texttt{Llama-3.1-8B-Instruct}) with the \enquote{LLM} role. This scenario has a single agent type \(\langle\) generalist, \enquote{LLM}\(\rangle\).
\item[II.\ Random roles.] All 48 agents run the generalist (\texttt{Llama-3.1-8B-Instruct}). We randomly assign each of the 16 roles in~\Cref{tab:models} to an agent. Each role is assigned \emph{exactly} 3 times per run: \(16 \times 3 = 48\).
\item[III.\ Random specialists.] Each of the 14 specialist LLMs powers three agents, and the generalist powers six (\(14 \times 3 + 6 = 48\)). Roles are assigned as in II, at random and independently of the LLM.
\item[IV.\ Matched specialists.] The LLM composition is the same as in III, but each agent holds the role that matches its LLM. Each of the 16 \(\langle\)LLM, role\(\rangle\) pairs in~\Cref{tab:models} is assigned to three agents at random. The generalist forms two pairs (\enquote{LLM} and \enquote{Human Participant}) and therefore powers six agents.
\end{description}
Assignments are drawn once per network realization and reused across all 20 statements. Note that in scenarios III and IV, we always put medical LLM into node 0 (see~\Cref{app:maximin-networks} for the motivation). \Cref{tab:scenarios} reports the number of unique agent types per scenario.

\begin{table*}[ht]
\centering
\begin{adjustbox}{max width=\linewidth}
\begin{tabular}{@{}lllc@{}}
\toprule
\textbf{Scenario} & \textbf{LLMs/Specializations} & \textbf{Social Roles}  & \textbf{(Unique) Agent types}  \\
\midrule
I.\ Base LLMs            & Generalist (48)                    & Only \enquote{LLM} (48)     & 1           \\
II.\ Random roles        & Generalist (48)                    & 16 Random roles (3)          & 16        \\
III.\ Random specialists & 14 specialists (3) + generalist (6) & 16 Random roles (3)        & \(\le 48\)  \\
IV.\ Matched specialists & 14 specialists (3) + generalist (6) & 16 \underline{Matched} roles (3)       & 16       \\
\bottomrule
\end{tabular}
\end{adjustbox}
\caption{\textbf{Composition of scenarios.} Every run has 48 agents. 
The columns show the LLM composition (\emph{specialization}) and \emph{social-role} assignments for each scenario. \emph{Unique agent types} counts the distinct \(\langle\)LLM, role\(\rangle\) combinations per network realization.
Numbers in parentheses specify how many LLMs or roles are assigned to agents: e.g., \enquote{14 specialists (3)} means that each of the 14 specialist LLMs powers three agents, and \enquote{16 random roles (3)} means that each role is held by three agents (\(14 \times 3 + 6 = 16 \times 3 = 48\)).
We fix the assignment of \(\langle\)LLM, role\(\rangle\) combinations for each network realization and reuse them for all 20 statements; each scenario has 640 runs (32 realizations \(\times\) 20 statements).
Scenarios I vs. II isolate the effect of the persona-style role assignment; \{I, II\} vs.\ \{III, IV\} isolate the effect of LLM heterogeneity (introduction of specialists); III vs. IV isolate the effect of role--specialization alignment.}
\label{tab:scenarios}
\end{table*}

%% file: sections/appendix/D_maximin.tex
\twocolumn
\section{Sampling with Maximin Criterion}
\label{app:maximin}
The design space of our experiments requires around $10^{5}$ simulations (statements $\times$ network realizations $\times$ scenarios $\times$ network structures), which is infeasible given our time and resource constraints. 
To reduce the number of simulations while ensuring that we select a diverse set of statements and network realizations, we use the \textbf{maximin criterion}~\cite{johnson1990maximin}.
This appendix describes the algorithm and its application.

\subsection{Maximin Criterion}

Given a candidate pool $\mathcal{X} = \{x_1, \ldots, x_N\} \subset \mathbb{R}^{d}$ of size $N$,  we select a subset $S^{\star}\subseteq \mathcal{X}$ of size $K$ whose minimum pairwise Euclidean distance is as large as possible:
\begin{align}
\label{eq:maximin}
S^{\star} &=\; \arg\max_{\substack{S \subseteq \mathcal{X} \\ |S| = K}} d_{\min}(S),
\\
d_{\min}(S) &:= \min_{\substack{x, y \,\in S \\ x \neq y}} \,\lVert x - y \rVert_{2}.
\end{align}
We use the farthest-point greedy algorithm of~\citet{gonzalez1985clustering} to perform the iterative selection in~\Cref{alg:maximin}.

\paragraph{Procedure.} We start by standardizing features of \(\mathcal{X}\) to zero mean and unit variance. We then either select the sample with the largest norm or provide an initial set \(S_0\).
At each iteration, we select the candidate from $\mathcal X$ with the greatest minimum distance to the already-selected set $S^{\star}$.
\begin{algorithm}[h]
\caption{Greedy maximin selection}
\label{alg:maximin}
\begin{algorithmic}[1]
\Require Pool $\mathcal{X} = \{x_1, \ldots, x_N\} \subset \mathbb{R}^{d}$,
         target size $K$, optional starting set $S_0$.
\Ensure  $S \subseteq \mathcal{X}$ approximating $S^{\star}$ in \eqref{eq:maximin}.
\State $\mathcal{X} \leftarrow \texttt{standardize}(\mathcal{X})$ 
\If{$S_{0}$ is non-empty}
    \State $S \gets S_{0}$
\Else
    \State $S \gets \{\, \arg\max_{x \in \mathcal{X}} \lVert x \rVert_{2} \,\}$
\EndIf
\While{$|S| < K$}
    \For{each $x \in \mathcal{X} \setminus S$}
        \State $d(x) \gets \min_{x' \in S} \, \lVert x - x' \rVert_{2}$
    \EndFor
    \State $S \gets S \cup \{\, \arg\max_{x \in \mathcal{X} \setminus S} d(x) \,\}$
\EndWhile
\State \Return $S$
\end{algorithmic}
\end{algorithm}

\subsection{Sampling Network Realizations with Maximin Criterion}
\label{app:maximin-networks}

\begin{table}[ht]
    \centering
\begin{adjustbox}{width=\columnwidth}
\begin{tabular}{rrccccrc}
\toprule
Seed & $p$ & $\bar{\ell}$ & $C$ & $\bar{k}$ & $\sigma_k$ & $k_0$ & Ext. \\
\midrule
\multicolumn{8}{l}{\textit{Erd\H{o}s--R\'enyi} ($n=48$)} \\
\midrule
2012155723 & 0.30 & 1.76 & 0.23 & 12.38 & 2.83 & 11 &  \\
2292123236 & 0.30 & 1.67 & 0.35 & 15.75 & 3.39 & 20 &  \\
  43019216 & 0.30 & 1.73 & 0.32 & 13.54 & 3.94 &  7 &  \\
1982018573 & 0.30 & 1.70 & 0.30 & 14.46 & 2.27 & 11 &  \\
3781092945 & 0.30 & 1.71 & 0.29 & 13.96 & 3.04 & 27 &  \\
2701326989 & 0.30 & 1.69 & 0.35 & 15.46 & 3.97 &  9 &  \\
3798279163 & 0.30 & 1.73 & 0.28 & 13.62 & 3.29 & 15 &  \\
 601535774 & 0.30 & 1.67 & 0.31 & 15.42 & 2.46 & 21 &  \\
 484618803 & 0.10 & 3.04 & 0.06 &  3.79 & 1.51 &  2 & \checkmark \\
1027860867 & 0.14 & 2.33 & 0.15 &  6.17 & 2.67 &  1 & \checkmark \\
1152158826 & 0.17 & 2.06 & 0.15 &  7.71 & 1.95 & 10 & \checkmark \\
2290866548 & 0.21 & 1.90 & 0.19 &  9.54 & 2.66 &  4 & \checkmark \\
3290684630 & 0.35 & 1.66 & 0.35 & 16.17 & 2.98 &  9 & \checkmark \\
 911947484 & 0.24 & 1.80 & 0.25 & 11.54 & 2.49 & 17 & \checkmark \\
2076510780 & 0.28 & 1.74 & 0.29 & 13.21 & 3.30 &  8 & \checkmark \\
2474802538 & 0.31 & 1.65 & 0.36 & 16.58 & 3.46 & 13 & \checkmark \\
\midrule
Seed & $k/\beta$ & $\bar{\ell}$ & $C$ & $\bar{k}$ & $\sigma_k$ & $k_0$ & Ext. \\
\midrule
\multicolumn{8}{l}{\textit{Watts--Strogatz} ($n=48$)} \\
\midrule
1769674741 &  8/\,.10 & 2.28 & 0.42 &  8.00 & 1.31 & 10 &  \\
 365860106 &  8/\,.10 & 2.62 & 0.57 &  8.00 & 0.50 &  7 &  \\
 395036430 &  8/\,.10 & 2.30 & 0.43 &  8.00 & 1.04 &  5 &  \\
3830640276 &  8/\,.10 & 2.35 & 0.47 &  8.00 & 0.65 &  9 &  \\
1093773836 &  8/\,.10 & 2.46 & 0.50 &  8.00 & 1.14 &  8 &  \\
1356044289 &  8/\,.10 & 2.45 & 0.51 &  8.00 & 0.79 &  6 &  \\
 851350617 &  8/\,.10 & 2.62 & 0.54 &  8.00 & 0.82 &  9 &  \\
4262934205 &  8/\,.10 & 2.28 & 0.41 &  8.00 & 0.98 &  8 &  \\
  41506191 & 16/\,.50 & 1.66 & 0.34 & 16.00 & 2.53 & 21 & \checkmark \\
 527896875 &  4/\,.03 & 4.54 & 0.45 &  4.00 & 0.35 &  4 & \checkmark \\
3175353684 & 14/\,.02 & 2.03 & 0.66 & 14.00 & 0.58 & 17 & \checkmark \\
3470624063 & 14/\,.50 & 1.72 & 0.32 & 14.00 & 2.52 & 10 & \checkmark \\
2378662268 & 12/\,.16 & 1.87 & 0.43 & 12.00 & 1.49 & 16 & \checkmark \\
 253806045 & 12/\,.01 & 2.47 & 0.68 & 12.00 & 0.00 & 12 & \checkmark \\
2181192797 &  6/\,.05 & 3.56 & 0.56 &  6.00 & 0.46 &  4 & \checkmark \\
 184331390 & 10/\,.09 & 2.20 & 0.53 & 10.00 & 0.96 & 13 & \checkmark \\
\bottomrule
\end{tabular}
\end{adjustbox}
\caption{\textbf{Overview of the structural features of the 16 networks selected per network structure} via Algorithm~\ref{alg:maximin}. These networks are generated using our network generators (ER and WS). Features: $(\bar{\ell})$ mean shortest-path length; $(C)$ global clustering coefficient; $(\bar{k})$ mean degree; $(\sigma_k)$ standard deviation of the degree distribution; $(k_0)$ degree of node $0$.
The initial eight realizations use fixed generator parameters (ER: $p=0.3$; WS: $k=8$, $\beta=0.1$). The eight realizations marked \emph{Ext.} form the extended set, which varies the generator parameters (ER: $p$; WS: $k$ and $\beta$) and uses the initial eight as the starting set $S_0$.}
    \label{tab:maximin-graphs}
\end{table}

We consider two network structures: Erd\H{o}s--R\'enyi (ER) and Watts--Strogatz (WS). For each structure, we run simulations across multiple network realizations to ensure the findings are not driven by a single sample (i.e., we want a structurally diverse set of realizations for each network structure).

\paragraph{Initial set.} We \emph{first} generate $N_{\text{pool}} = 1000$ candidate realizations by drawing independent seeds for the corresponding generator while holding the following parameters fixed:
$n = 48$ for both network structures, $p = 0.3$ for ER, and $k = 8$, $\beta = 0.1$ for WS.

For each sampled network realization, we compute five structural features:
\begin{enumerate}[itemsep=0pt, topsep=0pt]
    \item mean shortest-path length,
    \item global clustering coefficient,
    \item mean degree,
    \item standard deviation of the degree distribution,
    \item degree of node $0$, the node reserved for the medical LLM (\texttt{Llama3-Med42-8B}) in scenarios III--IV.
\end{enumerate}
We discard candidate network realizations with more than one connected component. 
The resulting per-structure feature matrices are passed to~\Cref{alg:maximin} with $K = 8$. 

\paragraph{Extended set.} 
In the initial set, realizations within a structure differ only by random seed, plus the two structures differ in density: for $p = 0.3$ the ER realizations have mean degree $\bar{k} \in [12.4, 15.8]$, and every WS realization has $\bar{k} = 8$ (\Cref{tab:maximin-graphs}). 
Since each LLM agent reads a summary of \emph{all} its neighbors' beliefs, an ER--WS difference could reflect how many neighbors an LLM agent has (i.e., not related to clustering or path length).

We select eight additional realizations per structure. Here, we vary the generator parameters (ER: $p$; WS: $k/\beta$) to select additional realizations, and we use the initial eight realizations as the starting sets $S_0$.
The extended set provides a wider density range within each structure (ER: $\bar{k} \in [3.8, 16.6]$; WS: $\bar{k} \in [4, 16]$), and the ranges overlap, so ER and WS realizations can be compared at matched mean degree.

As a result of the maximin sampling, we obtain 16 network realizations per network structure\footnote{The analysis in the main paper is based on all 32 network realizations.}. \Cref{tab:maximin-graphs} provides an overview of the selected realizations.

\subsection{Sampling Discussion Statements with Maximin Criterion}
\label{app:maximin-statements}
\begin{table*}[ht]
\centering
\begin{adjustbox}{width=\textwidth}
\begin{tabularx}{\linewidth}{r X c c c c}
\toprule
ID & Statement & $\hat{y}_{\text{doc}}$ & $a_{\text{doc}}$ & $a_{\text{other}}$ & $c$ \\
\midrule
\multicolumn{6}{l}{\textit{False statements} ($y^\star = 0$)} \\
\midrule
false\_0 & Silver is indicated for the treatment of keratosis pilaris. & 0 & 0.999 & 0.800 & 0.948 \\
false\_1 & Terbutaline is indicated for the treatment of cramps. & 1 & 0.078 & 0.067 & 0.966 \\
false\_2 & Isosorbide is indicated for the treatment of ischemia. & 1 & 0.012 & 0.867 & 0.024 \\
false\_3 & Atorvastatin is indicated for the treatment of aggression. & 0 & 1.000 & 0.000 & 0.022 \\
false\_4 & Manidipine is indicated for the treatment of septic. & 0 & 1.000 & 0.333 & 0.554 \\
false\_5 & VLTS-589 is indicated for the treatment of diabetic macular edema. & 1 & 0.383 & 0.867 & 0.386 \\
false\_6 & Nandrolone decanoate is indicated for the treatment of Duchenne muscular dystrophy. & 0 & 1.000 & 0.667 & 0.608 \\
false\_7 & Vitamin E is indicated for the treatment of pulmonary coccidioidomycosis. & 0 & 1.000 & 0.267 & 0.191 \\
false\_8 & Tetracycline is indicated for the treatment of throat. & 0 & 0.755 & 0.933 & 0.763 \\
false\_9 & Ethinylestradiol is indicated for the treatment of dry cough. & 0 & 1.000 & 0.067 & 0.301 \\
\midrule
\multicolumn{6}{l}{\textit{True statements} ($y^\star = 1$)} \\
\midrule
true\_0 & Methyl nicotinate is indicated for the treatment of aches. & 1 & 0.773 & 0.933 & 0.869 \\
true\_1 & Terbutaline is indicated for the treatment of asthma. & 0 & 0.000 & 0.133 & 0.986 \\
true\_2 & Warfarin is indicated for the treatment of pulmonary embolism. & 1 & 1.000 & 0.200 & 0.006 \\
true\_3 & Tetracycline is indicated for the treatment of Q fever. & 0 & 0.047 & 0.933 & 0.188 \\
true\_4 & Theophylline is indicated for the treatment of asthma. & 1 & 0.531 & 0.267 & 0.773 \\
true\_5 & Sildenafil is indicated for the treatment of pulmonary hypertension. & 1 & 1.000 & 0.533 & 0.510 \\
true\_6 & Polymyxin B is indicated for the treatment of infections of the urinary tract. & 0 & 0.438 & 0.133 & 0.571 \\
true\_7 & SNS-314 is indicated for the treatment of tumors. & 0 & 0.016 & 0.533 & 0.589 \\
true\_8 & Valomaciclovir is indicated for the treatment of viral infection. & 0 & 0.318 & 0.400 & 0.920 \\
true\_9 & Linezolid is indicated for the treatment of nosocomial pneumonia. & 1 & 0.703 & 0.200 & 0.307 \\
\bottomrule
\end{tabularx}
\end{adjustbox}
\caption{\textbf{List of 20 discussion statements selected via~\Cref{alg:maximin}} and their corresponding ground-truth label and four features. We divide the statements into false (\(y^\star=0\)) and true (\(y^\star=1\)) subtables. The four maximin features are the medical specialist’s predicted label \((\hat{y}_{\text{doc}})\), the medical specialist’s accuracy \((a_{\text{doc}})\), the mean accuracy of the other 14 LLMs \((a_{\text{other}})\), and the consensus score \((c)\) between the medical specialist and the \(15\) other \(\langle\)LLM, matched role\(\rangle\) pairs: \( 1 - \lvert p_{\text{doc}}(\textsc{true}\mid s) - \tilde{p}_{\text{other}}(\textsc{true}\mid s) \rvert\), where \(\tilde{p}_{\text{other}}\) is the median probability of the \(15\) other \(\langle\)LLMs, matched role\(\rangle\) pairs. The scores are computed based on the prompt from~\citet{savcisens2025trilemma}. \textit{Note}: We do not correct template-induced grammatical artifacts so that prompts remain uniformly formatted across conditions.}
\label{tab:maximin-statements}
\end{table*}

Our statements are drawn from the medical-indication data introduced in~\citet{savcisens2025trilemma}. 
We restrict selection to non-negated statements that reference real-world entities.\footnote{We exclude neither-valued statements with synthetic entities.} 
This dataset includes statements with labels derived from DrugBank~\cite{wishart2018drugbank}, which we treat as ground truth. An example of such a statement is \enquote{\textit{Heparin is indicated for the treatment of pulmonary embolism}.}

Before applying the maximin criterion, we compute zero-shot predictions for each statement \(s\) from the medical specialist (\texttt{Llama3-Med42-8B} with the matched role) and from 15 other \(\langle\)LLMs, matched role\(\rangle\) pairs, see \Cref{tab:models}. 
In addition, we define the following features for each statement~\(s\):
\begin{enumerate}[itemsep=1pt, topsep=1pt]
    \item The binary ground-truth label: \\\(y^\star \in \{ \textsc{false}, \textsc{true}\}\) or \(\{0,1\}\), respectively,
    \item Label predicted by a medical specialist,
    \item Medical specialist's accuracy, based on next-token probabilities: \(1 - \lvert y^\star - p_{\text{doc}}(\textsc{true}\mid s) \rvert\),
    \item Label-level (averaged) accuracy of other LLMs,
    \item Consensus between the medical specialist and other models \(= 1 - \lvert p_{\text{doc}}(\textsc{true}\mid s) - \tilde{p}_{\text{other}}(\textsc{true}\mid s) \rvert\), where \(\tilde{p}_{\text{other}}\) denotes the median probability assigned by the 15 other \(\langle\)LLM, matched role\(\rangle\) pairs.
\end{enumerate}
The first two features are binary; the remaining three are continuous in the range \([0,1]\). Note that  feature \(\#4\) is computed from the binary label predictions, while feature \(\#5\) compares probability distributions, so the two are not redundant.

Since we want to have an equal number of true and false statements, we add a constraint to~\Cref{alg:maximin} that enforces a \(K/2\) quota per ground-truth label (we set \(K=20\)). That is, we ensure that we select \(10\) true and \(10\) false statements. \Cref{tab:maximin-statements} displays selected statements with the corresponding features.

%% file: sections/appendix/F_influence.tex
\section{Details on Outgoing Influence}
\label{app:influence-details}

This section gives the full specification of the \emph{outgoing influence} introduced in \S\ref{sec:descriptive}.

Let \(\Delta\mathbf b_i^{(t)}=\mathbf b_i^{(t)}-\mathbf b_i^{(t-1)}\) be agent \(a_i\)'s per-round belief change (as in \Cref{eq:belief-change}), and let \(\mathcal N(i)\) be the set of \(a_i\)'s neighbors in the network \(G\).

Let \(\mathbf u_i^{(t)}\) be the direction in which LLM agent \(a_i\) updates its belief in round \(t\):
\begin{equation}
\mathbf u_i^{(t)}=\Delta\mathbf b_i^{(t)}\big/\lVert\Delta\mathbf b_i^{(t)}\rVert_2,\qquad \Delta\mathbf b_i^{(t)}\neq\mathbf 0.
\end{equation}
After \(a_i\) updates its belief in round \(t\), we compute how far the agents in a set \(S\) move along \(\mathbf u_i^{(t)}\) in the next round, \(t+1\):
\begin{equation}
\alpha_i^{(t)}(S)=\frac{1}{|S|}\sum_{j\in S}\bigl\langle\mathbf u_i^{(t)},\Delta\mathbf b_j^{(t+1)}\bigr\rangle.
\end{equation}
Further, we compute a weighted average of \(\alpha_i^{(t)}(S)\) over rounds, weighting each round by the size of \(a_i\)'s move:
\begin{equation}
\begin{aligned}
\pi_i(S)&=\sum_{t=1}^{T-1}w_i^{(t)}\,\alpha_i^{(t)}(S),\\
w_i^{(t)}&=\lVert\Delta\mathbf b_i^{(t)}\rVert_2\Big/\textstyle\sum_{s=1}^{T-1}\lVert\Delta\mathbf b_i^{(s)}\rVert_2.
\end{aligned}
\label{eq:influence-full}
\end{equation}
Rounds in which \(a_i\) does not move get zero weight.

When we set \(S=\mathcal N(i)\), we get \(\pi_i(\mathcal N(i))\): the average distance that \(a_i\)'s neighbors move along \(a_i\)'s direction in the round after \(a_i\) moves.

\paragraph{Null model.}
We compare \(a_i\)'s neighbors with its neighbors in degree-preserving rewirings of \(G\).
For each network realization, we draw \(20\) rewired networks \(\tilde G_1,\dots,\tilde G_{20}\) by double-edge swaps~\cite{maslov2002specificity}: two edges \((a,b)\) and \((c,d)\) are replaced by \((a,d)\) and \((c,b)\).
We reject swaps that would create a self-loop or a multi-edge.

Every agent keeps its original degree, so \(a_i\) has as many neighbors in each rewired network as in \(G\), with \(|\tilde{\mathcal N}_r(i)|=|\mathcal N(i)|\).
The null term averages over the rewirings:
\begin{equation}
\pi_i\bigl(\bar{\mathcal N}(i)\bigr):=\frac{1}{R}\sum_{r=1}^{R}\pi_i\bigl(\tilde{\mathcal N}_r(i)\bigr).
\end{equation}
Thus, \(\pi_i\bigl(\bar{\mathcal N}(i)\bigr)\) is the alignment with \(a_i\)'s moves expected from equally many agents in a degree-preserving rewired network \(\tilde{G}\).
This term accounts for alignment that arises by chance.

\paragraph{Outgoing Influence.}
The outgoing influence in \Cref{eq:influence} is the difference between observed alignment and random alignment: 
\begin{equation}
    \label{eq:outgoing-influence-full}
    \iota_i=\pi_i(\mathcal N(i))-\pi_i(\bar{\mathcal N}(i)).
\end{equation}
That is, \(\iota_i\) is a move-weighted average, over rounds, of how much farther \(a_i\)'s neighbors move in \(a_i\)'s direction than its random neighbors do.
\(\iota_i>0\) means that neighbors follow \(a_i\) beyond random drift, \(\iota_i\approx0\) that they do not, and \(\iota_i<0\) that the neighbors align with the agent's moves less than its rewired neighbors do.
\(\iota_i\) is undefined if \(a_i\) never moves in rounds \(1,\dots,T-1\); this occurs for no agent in our data.
Note that \(\iota_i\) measures association, not causation. 

\paragraph{Persistence.}
In~\Cref{app-sec:variance-decomp}, we describe the \emph{variance decomposition} method (intraclass correlation) that allows us to identify opinion-leader-like agents.

%% file: sections/appendix/E_descriptive_analysis.tex
\onecolumn
\section{Additional Details on Descriptive Analysis}
\label{app:descriptive}

This section expands on the methodology described in~\S\ref{sec:eval}.
Here, we use \texttt{R}'s mixed-model (\texttt{lme4}) formula notation~\cite{bates2015lmer}. A term \texttt{\detokenize{y ~ x + (1|g)}} fits a fixed effect for \texttt{x} and a random intercept for each level of grouping factor \texttt{g} (a separate baseline offset per level of \texttt{g}, which accounts for repeated measures within each level). 
Meanwhile, \texttt{\detokenize{(1|g1:g2)}} denotes a random intercept for each combination of levels of \texttt{\detokenize{g1}}  and \texttt{\detokenize{g2}}; it is used when \texttt{\detokenize{g2}}'s labels are only meaningful jointly with \texttt{\detokenize{g1}}. For instance, \enquote{network realization 3} refers to a different network under ER vs. WS.  
We provide the overview of the scenarios in~\Cref{tab:scenarios}.

Finally, \texttt{\detokenize{n-str}} stands for the \textit{network structure} (ER or WS) and \texttt{\detokenize{n-realization}} for the \textit{network realization}; each network structure has 16 realizations~(\Cref{tab:maximin-graphs}).
We use two identifiers for a network realization, depending on the analysis: \texttt{\detokenize{n-realization}} indexes a realization \textit{within} its network structure (values 1--16, reused across ER and WS) and \texttt{\detokenize{n-realization-id}} is a globally unique identifier assigned to each of the 32 realizations across both structures.
\texttt{run} identifies a simulation run and \texttt{agent} an agent (node) within a network realization.

\subsection{Linear Mixed-Effects Models}
\label{app:lmm}

The design of our experiments is fully crossed: every network realization and every statement appear under four scenarios (\(4\times32\times20=2{,}560\) runs; \(122{,}880\) agent rows, 48 agents per run). 
Agent-level outcomes (plasticity, directedness, outgoing influence) are fit on agent rows and consensus (and consensus change) on runs. 

\paragraph{Estimated marginal means.}
Estimated marginal means (EMMs), or least-squares means, are model-based predicted means adjusted for the other terms in the mixed model. 
To estimate the coefficients, EMMs, and confidence intervals, we fit the following linear mixed-effects model:
\begin{center}
\texttt{\detokenize{<var-name> ~ scenario * n-str + (1 | statement) + (1 | n-str:n-realization)}}.
\end{center}
The random-effects structure includes \texttt{\detokenize{(1|statement)}} to account for repeated measures across statements and \texttt{\detokenize{(1|n-str:n-realization)}} to capture variation due to different realizations of each network structure.
We provide an overview of results in~\Cref{tab:agg-results}. Additionally, we include the marginal estimates where the mixed-effects model is specified as:
\begin{center}
\texttt{\detokenize{<var-name> ~ scenario + (1 | statement) + (1 | n-str:n-realization)}}.
\end{center}

\subsection{Variance Decomposition} 
\label{app-sec:variance-decomp}
To assess whether some agents exert more influence than others, we fit a variance‑components model to the outgoing influence \(\iota_i\) (\Cref{app:influence-details}), separately for each scenario (\(30{,}720\) agent rows each), with an intercept-only fixed effect. The model formula is
\begin{center}
\small{
\texttt{\detokenize{out-influence ~ 1 + (1 | n-realization-id) + (1 | statement) + (1 | agent:n-realization-id)}}}.
\end{center}
It allows us to quantify the variance in \textit{outgoing influence} attributed to agent (LLM and social role) relative to network realization and statement.
For each random‑effects component, we report the intraclass correlation coefficient (ICC) as the proportion of total variance attributable to that component.
A higher agent‑level ICC suggests  persistent between-agent differences in how strongly neighbors align with agent \(a_i\)'s belief updates (opinion-leader-like agents).
A near-zero value indicates that outgoing influence is largely exchangeable across LLM agents. Low outgoing influence does not by itself identify a follower.

In~\Cref{fig:influence-panel}, we report \(95\%\) confidence intervals estimated via a parametric bootstrap (\(1{,}000\) resamples).

To test whether the agent‑level variance component is distinguishable from zero, we use an exact restricted likelihood ratio test~(\texttt{RLRsim};~\citealp{scheipl2008sizeap}), which accounts for the boundary problem in testing \(\sigma^2_{\text{agent}} = 0\).
The null distribution is simulated with \(10{,}000\) Monte Carlo draws.

\subsection{Contrast Analysis}
\label{app:contrast-analysis}

We define four planned contrasts on the four scenarios (I, II, III, and IV), each corresponding to a single‑degree‑of‑freedom effect on the outcome.  These four contrasts were specified a priori, before inspecting the fitted models. 

\paragraph{Contrast models.}
The EMM model has only random intercepts shared by all four scenarios (these would cancel in any scenario contrast). 
Here we fit a richer model that lets scenario effects vary by statement and by realization. For agent-level outcomes:
\begin{center}
\small{\texttt{\detokenize{<var-name> ~ scenario * n-str + (1 | statement) + (1 | scenario:statement) + (1 | n-str:statement) + (1 | n-realization-id) + (1 | scenario:n-realization-id) + (1 | run) + (1 | agent:n-realization-id:scenario)}}}
\end{center}
and for run-level outcomes (consensus change):
\begin{center}
\small{\texttt{\detokenize{<var-name> ~ scenario * n-str + (1 | statement) + (1 | scenario:statement) + (1 | n-realization-id) + (1 | scenario:n-realization-id)}}}
\end{center}
The \texttt{\detokenize{scenario:statement}} and \texttt{\detokenize{scenario:n-realization-id}} terms do not cancel in scenario contrasts: contrast precision is determined by the 20 statements and 32 realizations rather than by the number of LLM-agent rows from every simulation (Satterthwaite degrees of freedom \(\approx 60\)).
\texttt{\detokenize{(1 | run)}} and \texttt{\detokenize{(1 | agent:n-realization-id:scenario)}} absorb the clustering of the 48 agents within a run and the repeated measurement of the same \(\langle\)LLM, role\(\rangle\) assignment across the 20 statements (\Cref{tab:scenarios}).

\paragraph{Contrasts.} 
We define the contrast as \(c = \mathrm{EMM}_{1} - \mathrm{EMM}_2\), where  EMMs are based on the outcomes of two scenarios.
\Cref{tab:contrasts} specifies the contrast weights, and below we provide the contrast definitions:
\begin{description}[itemsep=0pt,topsep=0pt]
    \item[Role Effect:] Random Roles (II) vs. Baseline (I)
    \item[Specialization Effect:] Random Specialists (III) vs. Random Roles (II)
    \item[Role--specialization (alignment) effect:] Matched Specialists (IV) vs. Random Specialists (III)
    \item[Composition Effect:] 
    comparing the average of homogeneous scenarios (I and II) and the average of heterogeneous scenarios (III and IV).
\end{description}

\begin{table}[ht]
\centering
\begin{adjustbox}{width=\textwidth}
\begin{tabular}{rccccc}
\toprule
\textbf{Contrast} & Notation & \makecell{I. Baseline} & \makecell{II. Random\\Roles} & \makecell{III. Random\\Specialists} & \makecell{IV. Matched\\Specialists} \\
\midrule
Role effect           & I\(\to\)II & $-1$   & $+1$   & $0$    & $0$    \\
Specialization effect          & II\(\to\)III & $0$    & $-1$   & $+1$   & $0$    \\
Role--specialization alignment  & III\(\to\)IV & $0$    & $0$    & $-1$   & $+1$   \\
Composition effect         & \{I, II\}\(\to\)\{III,IV\} & $-0.5$ & $-0.5$ & $+0.5$ & $+0.5$ \\
\bottomrule
\end{tabular}
\end{adjustbox}
\caption{\textbf{Contrast weights for the four contrasts between scenarios.} Each row defines a linear combination of the four scenarios as a single degree-of-freedom contrast. Resulting effect sizes are visualized in~\Cref{fig:contrast-forest} and reported in~\Cref{tab:marginal-contrasts,tab:conditional-contrasts}.}
\label{tab:contrasts}
\end{table}
For each contrast above, we compute three quantities:
\begin{itemize}[itemsep=0pt,topsep=3pt]
    \item \textit{Marginal} contrast (\Cref{tab:marginal-contrasts}): the contrast averaged across both network structures, showing the overall effect of a manipulation.
    \item \textit{Conditional} contrast (\Cref{tab:conditional-contrasts}): the same linear combination estimated separately within each network structure (ER or WS).
    \item \textit{Interaction} contrast (the \textit{Between} rows in~\Cref{tab:conditional-contrasts}): the difference between the ER and WS conditional contrasts, \(c_{\text{ER}} - c_{\text{WS}}\). A non-significant interaction indicates that we do not detect differences between manipulation effects when comparing two network structures. 
\end{itemize}
For all contrasts, we report Cohen’s  \(d\), \(95\%\) confidence intervals, standard errors, and \(t\)-statistics.
These estimates are obtained with \texttt{emmeans}~\cite{russel2026emmeans} from the model-based variance--covariance matrix; no aggregation is applied before fitting.
\begin{figure*}[ht]
    \centering
    \includegraphics[width=\linewidth]{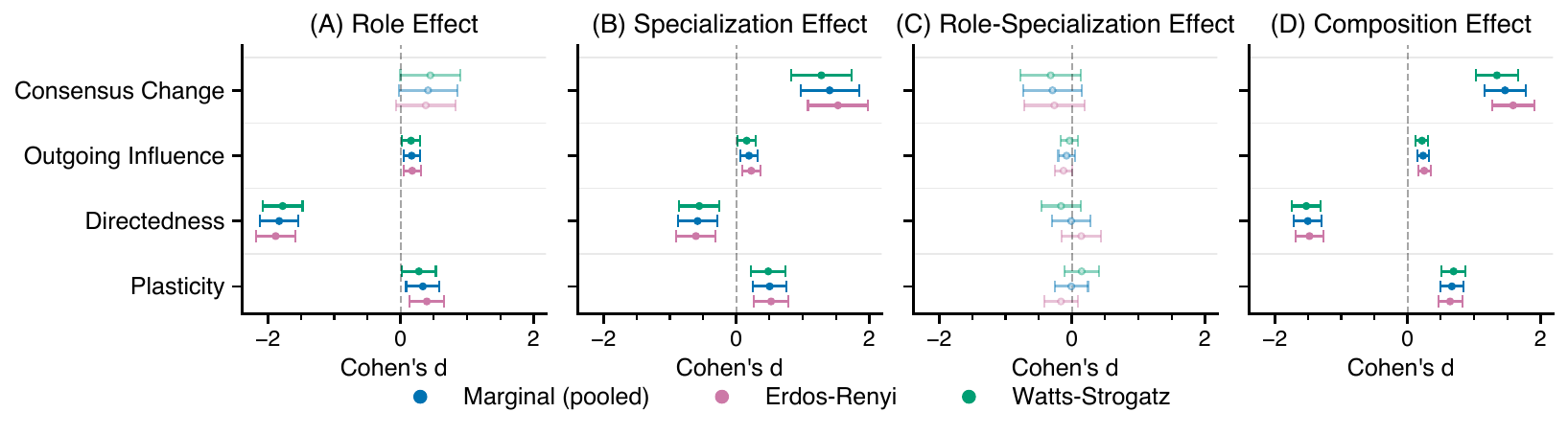}
    \caption{\textbf{Effect sizes for the four planned contrasts.} We provide marginal and conditional estimates for ER and WS network structures. The results are provided in~\Cref{tab:conditional-contrasts,tab:marginal-contrasts}.}
    \label{fig:contrast-forest}
\end{figure*}

\paragraph{Contrasts and Networks.}
We further test whether network structure modulates the planned contrasts using interaction contrasts \(\Delta c=c_{\mathrm{ER}}-c_{\mathrm{WS}}\) (the \emph{between} rows of \Cref{tab:conditional-contrasts}).
For these rows, we report Holm-adjusted \(p\)-values, applied separately within two levels: the agent level (3 outcomes \(\times\) 4 contrasts \(=12\) tests) and the run level (final consensus and consensus change \(\times\) 4 contrasts \(=8\) tests).
Final consensus is part of the run-level family but its interaction contrasts are not shown in \Cref{tab:conditional-contrasts}; they closely track those of consensus change (specialization: \(p=.011\) with Holm \(p=.067\); composition: \(p<.001\) with Holm \(p=.005\)).
We provide results in \Cref{tab:conditional-contrasts}.

\begin{figure*}[ht]
    \centering
    \includegraphics[width=\linewidth]{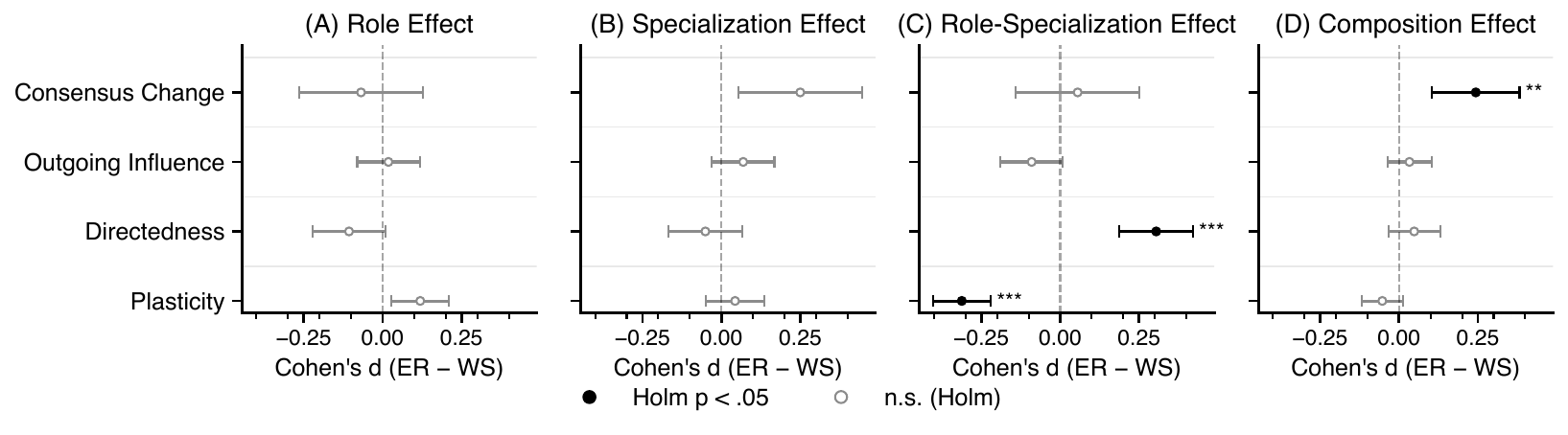}
    \caption{\textbf{Network structure effect on the four planned contrasts.} Each point is the ER\(-\)WS difference of a contrast (Cohen's \(d\), with \(95\%\) CI). Filled markers and stars mark Holm-adjusted significance (within the agent and run levels, 12 agent-level tests, 8 run-level tests). Estimates are in~\Cref{tab:conditional-contrasts}.}
    \label{fig:family-difference-forest}
\end{figure*}

Role--specialization alignment (III\(\to\)IV; \Cref{fig:family-difference-forest}C) lowers plasticity in ER (\(d=-0.17\)) but raises it in WS (\(d=0.15\); \(\Delta d=-0.31\), Holm \(p<.001\)).
Directedness shows the inverse pattern (\(d=0.14\) in ER vs.\ \(-0.16\) in WS; \(\Delta d=0.30\), Holm \(p<.001\)).
While the ER--WS difference is detectable, the alignment effects on plasticity and directedness are small and not significant within either structure: in \Cref{tab:conditional-contrasts}, the conditional effects of alignment on plasticity have \(p=.200\) (ER), and \(p=.257\) (WS).

The composition contrast (\Cref{fig:family-difference-forest}D) shows that the shift in consensus is larger in ER than in WS (\(\Delta d=0.243\), Holm \(p=.005\)); the specialization contrast (II\(\to\)III) points the same way (\(\Delta d=0.25\)) but does not survive adjustment (Holm \(p=.067\)).

Overall, network structure is a \emph{secondary} modulator. It changes the size of some effects, most clearly the specialist-driven consensus shift, but not the direction of any detectable effect.\footnote{With 16 realizations per network structure, we may lack the power to detect these effects.}

\paragraph{Computations and effect sizes.}
We compute the \textit{marginal}, \textit{conditional} and \textit{interaction} contrasts \(c\) and their \(95\%\) confidence intervals from the contrast models with \texttt{emmeans}~\citep{russel2026emmeans}, using Satterthwaite-approximated degrees of freedom (\texttt{lmerTest};~\citealp{kuznetsova2017lmer}).
For each contrast, we report Cohen's \(d\): the estimated marginal-mean (EMM) contrast \(c\) standardized by the \emph{total} standard deviation of the mixed-effects model,
\begin{equation}
\label{eq:cohen}
    d = \frac{c}{\hat{\sigma}_{\mathrm{total}}},
    \quad \text{with} \quad
    \hat{\sigma}_{\mathrm{total}}
      = \sqrt{\,\hat{\sigma}^{2}_{\epsilon}
              + \textstyle\sum_{g}\hat{\tau}^{2}_{g}\,},
\end{equation}
where we sum over the random-intercept variance components \(\hat{\tau}^{2}_{g}\) (e.g., all random-intercept components of the contrast model above, including the scenario-specific ones) and \(\hat{\sigma}^{2}_{\epsilon}\) is the residual variance~\citep{westfall2014power}. 
In \texttt{R}, it is computed as the square root of the
summed \texttt{VarCorr} components (valid for this purpose, since all random effects are intercept-only).
\textbf{Note} that because the components entering \(\hat{\sigma}_{\mathrm{total}}\) differ by observational level, \(d\) is comparable \emph{across scenarios within an outcome} but not directly across the agent and population levels.

The \textit{interaction} contrast (the \textit{Between} rows in~\Cref{tab:conditional-contrasts}) is the difference between the ER and WS conditional contrasts, \(\Delta c = c_{\mathrm{ER}} - c_{\mathrm{WS}}\), standardized by the same \(\hat{\sigma}_{\mathrm{total}}\) (values near zero indicate a network-invariant effect). 
The standard error for \(\Delta c\) (\texttt{SE}) is obtained from the variance--covariance matrix of the EMMs.

%% file: sections/appendix/G_calibration_check.tex
\onecolumn
\section{Beliefs and Confidence of LLM agents}
\label{app:calibration-check}

At round \(t=0\), we probe each LLM agent with the message template in~\Cref{fig:message}, omitting the neighborhood summary (\(\mathbf x_n^{(0)}=\emptyset\)). Therefore, an agent's round-0 belief reflects its prior on the statement together with any multiple-choice response bias.

\paragraph{Accuracy and calibration.}
\Cref{tab:round0-calibration} reports round-0 accuracy and calibration for each LLM under its matched role (scenario IV).
Accuracy ranges from \(0.50\) to \(0.85\), except for \texttt{UserLM-8B} (\(0.15\)), which abstains on \(80\%\) of statements.
\texttt{Code-Llama-3-8B} and \texttt{OpenMath2-Llama3.1-8B} label every statement as correct at round 0; on a balanced set of 10 true and 10 false statements, their accuracy of \(0.50\) reflects a response bias. We also report the Brier score and log loss.

\begin{table*}[h]
\centering
\begin{adjustbox}{max width=\textwidth, max totalheight=0.85\textheight}
\begin{tabular}{llccccccc}
\toprule
\textbf{Checkpoint} & \textbf{Role} & \textbf{Accuracy} & \textbf{Says true} & \textbf{Abstain} & \textbf{\(\bar p(y^\star)\)} & \textbf{mAP} & \textbf{Brier} & \textbf{Log loss} \\
\midrule
\texttt{lightblue/suzume-llama-3-8B-multilingual} & Language Mediator & 0.85 & 0.65 & 0.00 & 0.55 & 0.84 & 0.13 & 0.69 \\
\texttt{fdtn-ai/Foundation-Sec-8B-Instruct} & Cybersecurity Analyst & 0.80 & 0.40 & 0.00 & 0.78 & 0.91 & 0.10 & 0.79 \\
\texttt{meta-llama/Llama-3.1-8B-Instruct} & Human Participant & 0.70 & 0.40 & 0.00 & 0.63 & 0.74 & 0.17 & 1.01 \\
\texttt{meta-llama/Llama-3.1-8B-Instruct} & LLM & 0.70 & 0.40 & 0.00 & 0.61 & 0.72 & 0.18 & 1.04 \\
\texttt{m42-health/Llama3-Med42-8B} & Clinical Physician & 0.70 & 0.50 & 0.00 & 0.65 & 0.70 & 0.17 & 1.42 \\
\texttt{OpenSciLM/Llama-3.1\_OpenScholar-8B} & Academic Scholar & 0.70 & 0.40 & 0.00 & 0.48 & 0.72 & 0.16 & 0.79 \\
\texttt{NousResearch/Hermes-3-Llama-3.1-8B} & Strategic Planner & 0.65 & 0.25 & 0.00 & 0.65 & 0.79 & 0.17 & 1.09 \\
\texttt{FrancescoPeriti/Llama3Dictionary-merge} & Lexicographer & 0.65 & 0.45 & 0.00 & 0.51 & 0.75 & 0.15 & 0.70 \\
\texttt{USTC-KnowledgeComputingLab/Llama3-KALE-LM-Chem-1.5-8B} & Chemist & 0.60 & 0.50 & 0.00 & 0.48 & 0.68 & 0.16 & 0.79 \\
\texttt{HumanLLMs/Human-Like-LLama3-8B-Instruct} & Assistant & 0.55 & 0.55 & 0.00 & 0.59 & 0.73 & 0.20 & 0.95 \\
\texttt{ContactDoctor/Bio-Medical-Llama-3-8B} & Biomedical Researcher & 0.55 & 0.20 & 0.10 & 0.47 & 0.78 & 0.20 & 1.20 \\
\texttt{vicgalle/Roleplay-Llama-3-8B} & Storyteller & 0.55 & 0.45 & 0.00 & 0.59 & 0.73 & 0.20 & 0.99 \\
\texttt{ajibawa-2023/Code-Llama-3-8B} & Software Engineer & 0.50 & 1.00 & 0.00 & 0.50 & 0.82 & 0.21 & 0.93 \\
\texttt{mukaj/Llama-3.1-Hawkish-8B} & Financial Analyst & 0.50 & 0.40 & 0.00 & 0.51 & 0.65 & 0.19 & 0.87 \\
\texttt{nvidia/OpenMath2-Llama3.1-8B} & Mathematician & 0.50 & 1.00 & 0.00 & 0.49 & 0.68 & 0.18 & 0.79 \\
\texttt{microsoft/UserLM-8b} & Online Friend & 0.15 & 0.20 & 0.80 & 0.26 & 0.65 & 0.30 & 1.45 \\
\bottomrule
\end{tabular}
\end{adjustbox}
\caption{\textbf{Round-0 accuracy per LLM checkpoint} (scenario IV, matched roles; \(n=20\) statements per row).  \textit{Says true}: share of statements labelled \textit{correct} (the set is 10 true and 10 false); \textit{Abstain}: share labelled \textit{neither}; \(\bar p(y)\): mean probability on the DrugBank-derived label; \textit{mAP}: mean average precision of the true and false classes (threshold-free, chance \(0.50\)); Brier: squared error averaged over the three options (uniform guess \(0.22\)). Per-checkpoint ECE is omitted (20 items); pooled across checkpoints, ECE \(= 0.16\). Statements were selected to enrich disagreement with the medical LLM (Appendix~\ref{app:maximin}), so these values describe the priors agents bring into the interaction, not checkpoint capability.}
\label{tab:round0-calibration}
\end{table*}

\paragraph{Pooled calibration by scenario.}
Pooled over all cells of a scenario (\Cref{tab:round0-ece}), every agent population is overconfident: mean confidence exceeds accuracy in all four scenarios.
However, populations with specialist LLMs start less confident (III: \(0.72\) vs.\ I: \(0.81\) and II: \(0.83\)) and better calibrated (ECE \(0.12\)--\(0.16\) vs.\ \(0.19\)--\(0.22\)) than generalist populations.

\begin{table}[h]
\centering
\small
\begin{tabular}{lccccc}
\toprule
Scenario & Cells & Accuracy & Confidence (\(\pm\) SD) & mAP & ECE \\
\midrule
I. Base LLMs & 20 & 0.70 & 0.81 \(\pm\) 0.19 & 0.72 & 0.22 \\
II. Random roles & 320 & 0.64 & 0.83 \(\pm\) 0.17 & 0.71 & 0.19 \\
III. Random specialists & 4{,}540 & 0.60 & 0.72 \(\pm\) 0.19 & 0.68 & 0.12 \\
IV. Specialists (matched) & 320 & 0.60 & 0.72 \(\pm\) 0.19 & 0.67 & 0.16 \\
\bottomrule
\end{tabular}
\caption{\textbf{Pooled round-0 calibration by scenario.}
A cell is a unique (LLM, role, statement) combination; all runs of a cell receive the same round-0 prompt. \emph{Accuracy}: share of cells whose top belief label matches the ground truth. \emph{Confidence}: mean probability of the chosen option, with standard deviation across cells. \emph{mAP}: mean average precision of the true and false classes, pooled over all cells of the scenario. \emph{ECE}: expected calibration error with $10$ equal-width confidence bins (lower is better). Scenario I has only $20$ cells, so its ECE estimate is noisy.}
\label{tab:round0-ece}
\end{table}
 
\subsection{Are the specialist effects a confidence artifact?}
\label{app:round0-confidence}
 
Specialist LLMs start less confident than the generalist (mean top-label probability \(0.72\) in III vs.\ \(0.83\) in II), so \textit{they might revise more only because they are less certain.}
We refit the mixed-effects models of \Cref{app:descriptive} with identical random effects, once without and once with round-0 confidence as a centered covariate (\Cref{tab:round0-adjusted}).
For agent-level outcomes, an agent's round-0 confidence is the probability of its top label at round 0, taken from its (scenario, LLM, role, statement) cell (\Cref{tab:round0-ece}).
For run-level consensus change, we use the run's mean round-0 confidence, the unweighted average over its 48 agents.
In both cases, the covariate is centered on its overall mean and added as a fixed effect to the contrast models of \Cref{app:contrast-analysis}.
 
Less confident LLM agents do revise more (plasticity slope \(-0.16\), \(p<.001\)); 
however, adjusting for confidence reduces the estimated specialist (II\(\to\)III) plasticity contrast by about \(29\%\) (from \(0.056\) to \(0.040\)). 
The specialist effect (II\(\to\)III) on directedness becomes slightly stronger after adjustment.
More importantly, confidence \textit{does not predict consensus change} (\(p=.21\)).
The role and alignment contrasts are unaffected.

Among LLM agents whose round-0 confidence exceeds \(0.95\), those in generalist populations barely revise (mean plasticity \(0.003\) in II), whereas those in specialist populations still do (\(0.047\) in III).
\textbf{If lower confidence alone drove the specialist effect (II\(\to\)III), equally confident agents would revise about equally in both populations.}
Adjusting for initial confidence reduces the specialist plasticity contrast by about \(29\%\), but the remaining effect and the effects on directedness and consensus change persist after adjustment; the specialist effects are not explained exclusively by specialists' lower initial confidence.

\begin{table*}[t]
\centering
\small
\resizebox{\textwidth}{!}{%
\begin{tabular}{llccccc}
\toprule
 & & \multicolumn{2}{c}{Unadjusted} & \multicolumn{3}{c}{Adjusted for round-0 confidence} \\
\cmidrule(lr){3-4}\cmidrule(lr){5-7}
Outcome & Contrast & Estimate [95\% CI] & \(p\) & Estimate [95\% CI] & \(p\) & \(d\) \\
\midrule
Plasticity & Role (I\(\to\)II)            & 0.037 [0.009, 0.066] & .010 & 0.041 [0.014, 0.067] & .003 & 0.39 \\
           & Specialization (II\(\to\)III) & 0.056 [0.028, 0.084] & \(<.001\) & 0.040 [0.013, 0.066] & .004 & 0.38 \\
           & Alignment (III\(\to\)IV)     & \(-0.001\) [\(-0.029\), 0.027] & .940 & \(-0.001\) [\(-0.028\), 0.025] & .912 & \(-0.01\) \\
           & Confidence slope             & & & \(-0.159\) & \(<.001\) & \\
\midrule
Directedness & Role (I\(\to\)II)            & \(-0.494\) [\(-0.574\), \(-0.415\)] & \(<.001\) & \(-0.488\) [\(-0.571\), \(-0.406\)] & \(<.001\) & \(-1.76\) \\
             & Specialization (II\(\to\)III) & \(-0.158\) [\(-0.237\), \(-0.078\)] & \(<.001\) & \(-0.187\) [\(-0.269\), \(-0.105\)] & \(<.001\) & \(-0.68\) \\
             & Alignment (III\(\to\)IV)     & \(-0.003\) [\(-0.083\), 0.076] & .937 & \(-0.004\) [\(-0.086\), 0.078] & .925 & \(-0.01\) \\
             & Confidence slope             & & & \(-0.286\) & \(<.001\) & \\
\midrule
Consensus change & Role (I\(\to\)II)            & 0.042 [\(-0.003\), 0.086] & .066 & 0.044 [\(-0.002\), 0.089] & .059 & 0.44 \\
                 & Specialization (II\(\to\)III) & 0.142 [0.098, 0.187] & \(<.001\) & 0.133 [0.085, 0.180] & \(<.001\) & 1.33 \\
                 & Alignment (III\(\to\)IV)     & \(-0.030\) [\(-0.074\), 0.015] & .190 & \(-0.030\) [\(-0.075\), 0.016] & .194 & \(-0.30\) \\
                 & Confidence slope             & & & \(-0.092\) & .214 & \\
\bottomrule
\end{tabular}}
\caption{\textbf{Scenario contrasts with and without adjustment for round-0 confidence.} Differences in estimated marginal means (raw units), averaged over network structures; Satterthwaite degrees of freedom. The adjusted models evaluate EMMs at the mean of the centered covariate. \(d\): Cohen's \(d\) of the adjusted contrast (total-SD standardization, \Cref{app:descriptive}).
For consensus change, the covariate is the run's mean round-0 confidence over its \(48\) agents. The relative difference between unadjusted and adjusted estimates is attenuation under adjustment.
}
\label{tab:round0-adjusted}
\end{table*}

%% file: sections/appendix/H_surrogate_models.tex
\twocolumn
\section{Surrogate Model Details}
\label{app:dynamics}

The surrogate analysis evaluates whether the belief dynamics observed in \texttt{CoevolveSim} can be reproduced using four transition models that condition on different sources of information.

\subsection{Feature Construction}

\begin{table}[h]
\centering
\begin{adjustbox}{width=\linewidth}
\begin{tabular}{cll}
\toprule
\textbf{Model} & \textbf{Features} $\phi_i^{(t)}$ & \textbf{Mechanism} \\
\midrule
\textsc{M1} & $B_i^{(t)}$ & Individual persistence \\
\textsc{M2} & $B_i^{(t)}, \varGamma^{(t)}$ & Global belief composition \\
\textsc{M3} & $B_i^{(t)}, N_i^{(t)}$ & Local belief composition \\
\textsc{M4} & $B_i^{(t)}, N_i^{(t)}, M_i, R_i$ & Local influence + identity \\
\bottomrule
\end{tabular}%
\end{adjustbox}
\caption{
\textbf{Feature sets for the four surrogate models.} $\varGamma^{(t)}$ denotes the population-level belief distribution at round $t$, $N_i^{(t)}$ denotes the belief distribution within agent $i$'s neighborhood, $M_i$ denotes the underlying LLM, and $R_i$ denotes its assigned social role.
}
\label{tab:surrogate_models}
\end{table}

Let $B_i^{(r,t)} \in \{1,2,3\}$ denote the discretized belief of agent $a_i$ in run $r$ at round $t$, where $1$, $2$, and $3$ correspond to \textit{correct}, \textit{incorrect}, and
\textit{neither}, respectively. All models condition on the agent's current belief state $B_i^{(r,t)}$. Additional features are defined as follows.

\paragraph{Global belief composition.}
For run $r$ and round $t$, we define
\begin{equation}
\varGamma^{(r,t)}
=
\left(
f_{1}^{(r,t)},
f_{2}^{(r,t)},
f_{3}^{(r,t)}
\right),
\end{equation}
where $f_c^{(r,t)}$ denotes the fraction of agents holding belief state $c$.

\paragraph{Neighborhood belief composition.}
For agent $a_i$, we define
\begin{equation}
N_i^{(r,t)}
=
\left(
n_{i,1}^{(r,t)},
n_{i,2}^{(r,t)},
n_{i,3}^{(r,t)}
\right),
\end{equation}
where $n_{i,c}^{(r,t)}$ denotes the fraction of neighbors in $\mathcal N(i)$ holding belief state $c$.

\paragraph{Identity features.}
For model M4, we additionally include $M_i$ and $R_i$, the underlying LLM and social role assigned to agent $a_i$, respectively.

\subsection{Model Specification}

Each surrogate estimates a conditional distribution over next-round beliefs,
\begin{equation}
\Pr\!\left(
B_i^{(t+1)} = c
\mid
\boldsymbol\phi_i^{(t)}
\right),\;
c \in \{1,2,3\},
\end{equation}
where $\boldsymbol\phi_i^{(t)}$ is a model-specific feature vector. Models are fit separately for each experimental scenario and network structure.

\paragraph{M1: Empirical persistence baseline.}
M1 estimates one-step transition probabilities directly from observed transition frequencies:
\begin{equation}
\Pr\!\left(
B_i^{(t+1)} = c
\mid
B_i^{(t)}
\right).
\end{equation}

\paragraph{M2: Global belief composition.}
M2 conditions on the agent's current belief and the global belief composition:
\begin{equation}
\Pr\!\left(
B_i^{(t+1)} = c
\mid
B_i^{(t)},
\varGamma^{(t)}
\right).
\end{equation}

\paragraph{M3: Local belief composition.}
M3 conditions on the agent's current belief and local neighborhood (belief) composition:
\begin{equation}
\Pr\!\left(
B_i^{(t+1)} = c
\mid
B_i^{(t)},
N_i^{(t)}
\right).
\end{equation}

\paragraph{M4: Local belief composition + identity.}
M4 augments M3 with agent-level identity features:
\begin{equation}
\Pr\!\left(
B_i^{(t+1)} = c
\mid
B_i^{(t)},
N_i^{(t)},
M_i,
R_i
\right).
\end{equation}

\subsection{Fitting Procedure}

For each network structure and scenario (over \(320\) runs with \(16\) network realizations \(\times\) \(20\) statements, and \(48\) agents and \(10\) rounds per run, we get \(153{,}600\) one-step transitions), we construct transition-level examples from the observed belief trajectories of \texttt{CoevolveSim}.
We evaluate the fits with 5-fold cross-validation at the run level: each fold holds out 64 runs and fits on the remaining 256, the same fold assignment is used for all four surrogates, and each run is scored exactly once by a surrogate that never saw it.
Held-out runs share statements and network realizations with the training runs; the evaluation therefore measures generalization to new runs of known statements and networks, not to unseen statements or unseen networks.

M1 estimates transition probabilities directly from observed transition counts.
M2--M4 are multinomial logistic regressions with categorical inputs (current local or global belief composition, and agent's \(a_i\)'s LLM and role) one-hot encoded and composition fractions standardized.
There is no hyperparameter tuning.

\subsection{Evaluation}

We evaluate the surrogate models using both held-out prediction and behavioral rollouts.

\paragraph{Held-out one-step prediction.}
For each held-out run, the fitted surrogate predicts $\widehat{B}_i^{(T)}$ from the observed features $\boldsymbol{\phi}_i^{(T-1)}$. We evaluate predictive performance using the multiclass Matthews correlation coefficient (MCC) between $\widehat{B}_i^{(T)}$ and the observed $B_i^{(T)}$, pooled over agents and held-out runs. We refer to this measure as final-state MCC.

\paragraph{Rollout evaluation.}
For each held-out run, we initialize the surrogate from the empirical round-\(0\) belief state and generate rounds \(1,\dots,T\) by sampling each agent's next belief from the fitted transition probabilities (one stochastic rollout per held-out run, fixed seeds).
Rollouts are closed-loop: at every step, the global composition \(\varGamma^{(t)}\) and the neighborhood composition \(N_i^{(t)}\) are recomputed from the \emph{simulated} beliefs, while the network and the agents' identity features (\(M_i, R_i\)) are held fixed.
In contrast, final-state MCC is an open-loop, one-step prediction from the \emph{observed} round-\((T-1)\) state.

We compare empirical and surrogate trajectories using \textit{consensus fidelity}, which measures agreement between the observed and simulated consensus levels across rounds. For the held-out set of runs $\mathcal R_{\mathrm{test}}$, we define
\begin{equation}
\begin{split}
\mathrm{CF}
  &= \frac{1}{|\mathcal R_{\mathrm{test}}|}
     \sum_{r\in\mathcal R_{\mathrm{test}}}
     \frac{1}{T}\sum_{t=1}^{T} \\
  &\quad \Bigl(1 - \bigl|\mathrm{Cons}_{\mathrm{emp}}^{(r,t)}
                 - \mathrm{Cons}_{\mathrm{sim}}^{(r,t)}\bigr|\Bigr).
\end{split}
\end{equation}

Higher consensus fidelity indicates that a surrogate more closely reproduces the observed consensus trajectory.
We provide an overview of the results in~\Cref{tab:final-mcc,tab:consensus-fidelity}.
Confidence intervals are \(95\%\) percentile intervals from \(1{,}000\) bootstrap resamples of the 16 network realizations (each a cluster of 20 runs).

\subsection{Additional Results}

\begin{figure}[ht]
    \centering
    \includegraphics[width=\linewidth]{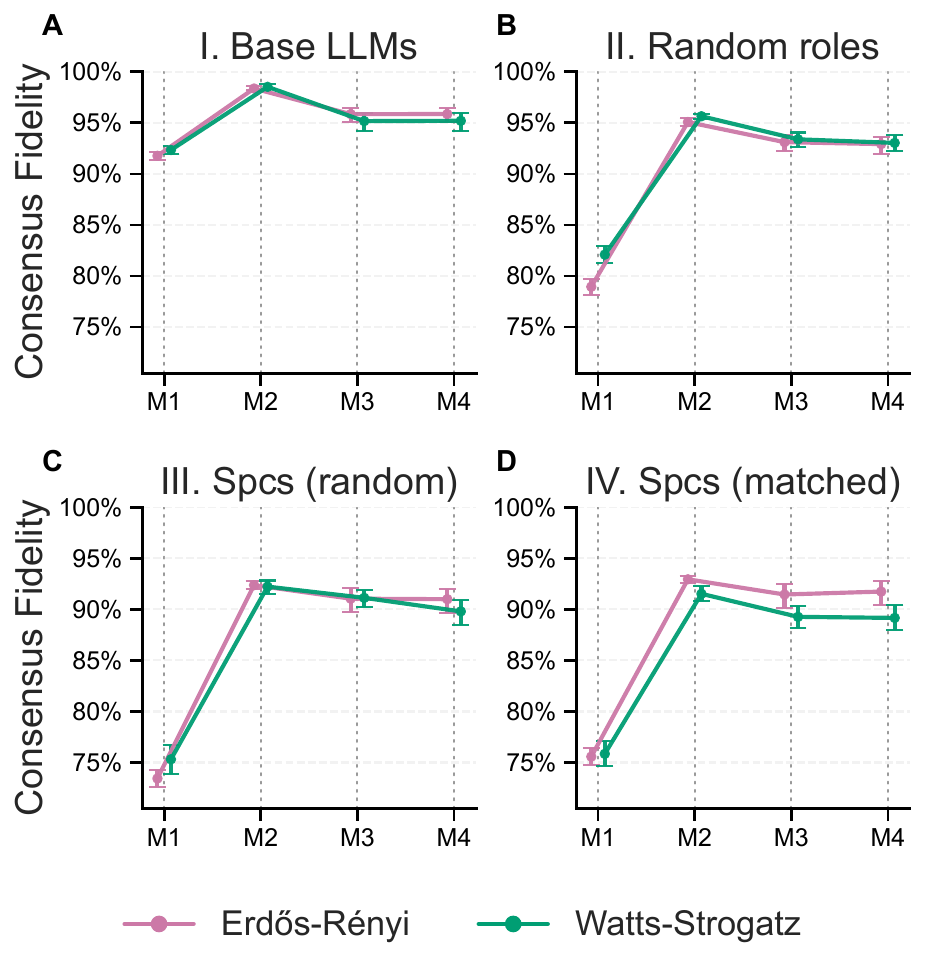}
    \caption{\textbf{Consensus fidelity across scenarios and surrogate models.} 
    Consensus fidelity measures agreement between empirical and surrogate consensus trajectories, averaged over rounds and held-out runs. \textit{Spcs} stands for \textit{specialists}. Panels show (A) Base LLMs, (B) Random roles, (C) Specialists with random roles, and (D) Specialists with matched roles. Points show mean estimates with $95\%$ bootstrap confidence intervals, obtained by resampling network realizations (\(1{,}000\) resamples), for Erd\H{o}s--R\'enyi (ER) and Watts--Strogatz (WS) networks. Persistence alone (M1) yields the lowest consensus fidelity in every scenario, whereas adding population-level belief composition (M2) yields the highest. Detailed results are provided in~\Cref{tab:consensus-fidelity}.
}
    \label{fig:pooled-consensus}
\end{figure}

In~\Cref{fig:pooled-consensus}, we visualize the consensus fidelity across surrogate models and scenarios. These results are also provided in~\Cref{tab:consensus-fidelity}.
As mentioned in~\S\ref{sec:surrogate_results}, adding population-level belief composition (M2) provides the best consensus fidelity across all scenarios.

%% file: sections/appendix/Y_tables.tex
\onecolumn
\section{Tables}
In this section, we provide detailed tables for 
\begin{enumerate}[itemsep=0pt, topsep=3pt]
    \item Estimated marginal means (EMMs) in~\Cref{tab:agg-results},
    \item Variance decomposition for the outgoing influence in~\Cref{tab:var-decomp},
    \item Contrast analysis: marginal effect sizes in~\Cref{tab:marginal-contrasts} and conditional effects by network structure in~\Cref{tab:conditional-contrasts},
    \item Surrogate analysis: final-state Matthews Correlation Coefficient (final-state MCC) in~\Cref{tab:final-mcc} and consensus fidelity in~\Cref{tab:consensus-fidelity}.
\end{enumerate}

\begin{table}
\centering
\begin{adjustbox}{max width=0.95\textwidth, max totalheight=0.80\textheight}
\begin{tabular}{llccccccc}
\toprule
 &  &  &  & \makecell{\(95\%\) CI} & \makecell{\(95\%\) CI}  &  & \\
Outcome & Scenario & Network & Estimate & lower & upper & SE & df \\
\midrule
\multirow[c]{12}{*}{Plasticity} & \multirow[c]{3}{*}{I. Base LLMs} & ER & 0.027 & 0.002 & 0.052 & 0.012 & 20 \\
 &  & WS & 0.026 & 0.001 & 0.052 & 0.012 & 20 \\
 &  & Marginal & 0.027 & 0.001 & 0.052 & 0.012 & 20 \\
\cline{2-8}
 & \multirow[c]{3}{*}{II. Random roles} & ER & 0.071 & 0.046 & 0.097 & 0.012 & 20 \\
 &  & WS & 0.057 & 0.032 & 0.082 & 0.012 & 20 \\
 &  & Marginal & 0.064 & 0.039 & 0.089 & 0.012 & 20 \\
\cline{2-8}
 & \multirow[c]{3}{*}{III. Random specialists} & ER & 0.130 & 0.104 & 0.155 & 0.012 & 20 \\
 &  & WS & 0.111 & 0.085 & 0.136 & 0.012 & 20 \\
 &  & Marginal & 0.120 & 0.095 & 0.146 & 0.012 & 20 \\
\cline{2-8}
 & \multirow[c]{3}{*}{IV. Specialists (matched)} & ER & 0.111 & 0.086 & 0.137 & 0.012 & 20 \\
 &  & WS & 0.127 & 0.102 & 0.153 & 0.012 & 20 \\
 &  & Marginal & 0.119 & 0.094 & 0.144 & 0.012 & 20 \\
\cline{1-8} \cline{2-8}
\multirow[c]{12}{*}{Directedness} & \multirow[c]{3}{*}{I. Base LLMs} & ER & 0.97 & 0.91 & 1.03 & 0.028 & 22 \\
 &  & WS & 0.97 & 0.91 & 1.03 & 0.028 & 22 \\
 &  & Marginal & 0.97 & 0.91 & 1.03 & 0.028 & 21 \\
\cline{2-8}
 & \multirow[c]{3}{*}{II. Random roles} & ER & 0.46 & 0.40 & 0.52 & 0.028 & 22 \\
 &  & WS & 0.49 & 0.43 & 0.55 & 0.028 & 22 \\
 &  & Marginal & 0.47 & 0.42 & 0.53 & 0.028 & 21 \\
\cline{2-8}
 & \multirow[c]{3}{*}{III. Random specialists} & ER & 0.29 & 0.23 & 0.35 & 0.028 & 22 \\
 &  & WS & 0.34 & 0.28 & 0.40 & 0.028 & 22 \\
 &  & Marginal & 0.32 & 0.26 & 0.37 & 0.028 & 21 \\
\cline{2-8}
 & \multirow[c]{3}{*}{IV. Specialists (matched)} & ER & 0.33 & 0.27 & 0.39 & 0.028 & 22 \\
 &  & WS & 0.29 & 0.24 & 0.35 & 0.028 & 22 \\
 &  & Marginal & 0.31 & 0.25 & 0.37 & 0.028 & 21 \\
\cline{1-8} \cline{2-8}
\multirow[c]{12}{*}{Outgoing Influence} & \multirow[c]{3}{*}{I. Base LLMs} & ER & -0.0002 & -0.0060 & 0.0056 & 0.003 & 42 \\
 &  & WS & -0.0022 & -0.0080 & 0.0036 & 0.003 & 42 \\
 &  & Marginal & -0.0012 & -0.0064 & 0.0040 & 0.003 & 32 \\
\cline{2-8}
 & \multirow[c]{3}{*}{II. Random roles} & ER & 0.0118 & 0.0060 & 0.0175 & 0.003 & 42 \\
 &  & WS & 0.0085 & 0.0027 & 0.0143 & 0.003 & 42 \\
 &  & Marginal & 0.0101 & 0.0049 & 0.0154 & 0.003 & 32 \\
\cline{2-8}
 & \multirow[c]{3}{*}{III. Random specialists} & ER & 0.0272 & 0.0214 & 0.0330 & 0.003 & 42 \\
 &  & WS & 0.0192 & 0.0134 & 0.0250 & 0.003 & 42 \\
 &  & Marginal & 0.0232 & 0.0179 & 0.0284 & 0.003 & 32 \\
\cline{2-8}
 & \multirow[c]{3}{*}{IV. Specialists (matched)} & ER & 0.0185 & 0.0127 & 0.0243 & 0.003 & 42 \\
 &  & WS & 0.0167 & 0.0109 & 0.0225 & 0.003 & 42 \\
 &  & Marginal & 0.0176 & 0.0124 & 0.0228 & 0.003 & 32 \\
\cline{1-8} \cline{2-8}
\multirow[c]{12}{*}{Consensus Change} & \multirow[c]{3}{*}{I. Base LLMs} & ER & -0.01 & -0.03 & 0.02 & 0.012 & 29 \\
 &  & WS & -0.01 & -0.03 & 0.02 & 0.012 & 29 \\
 &  & Marginal & -0.01 & -0.03 & 0.02 & 0.012 & 24 \\
\cline{2-8}
 & \multirow[c]{3}{*}{II. Random roles} & ER & 0.03 & 0.01 & 0.06 & 0.012 & 29 \\
 &  & WS & 0.04 & 0.01 & 0.06 & 0.012 & 29 \\
 &  & Marginal & 0.03 & 0.01 & 0.06 & 0.012 & 24 \\
\cline{2-8}
 & \multirow[c]{3}{*}{III. Random specialists} & ER & 0.19 & 0.16 & 0.21 & 0.012 & 29 \\
 &  & WS & 0.17 & 0.14 & 0.19 & 0.012 & 29 \\
 &  & Marginal & 0.18 & 0.15 & 0.20 & 0.012 & 24 \\
\cline{2-8}
 & \multirow[c]{3}{*}{IV. Specialists (matched)} & ER & 0.16 & 0.13 & 0.18 & 0.012 & 29 \\
 &  & WS & 0.13 & 0.11 & 0.16 & 0.012 & 29 \\
 &  & Marginal & 0.15 & 0.12 & 0.17 & 0.012 & 24 \\
\cline{1-8}
\bottomrule
\end{tabular}
\end{adjustbox}
\caption{\textbf{Estimated marginal means (EMMs) of agent- and population-level outcomes} by scenario and network structure (\Cref{app:descriptive}). EMMs with \(95\%\) confidence intervals are estimated with the linear mixed-effects models, reported per network structure (ER, WS) and marginally (averaged over both structures with equal weights). Agent-level outcomes (plasticity, directedness, outgoing influence) are fit with random intercepts for statement and network realization, while consensus change is fit at the run level; \texttt{df} denotes Satterthwaite-approximated degrees of freedom (which is why marginal and per-network rows differ). \texttt{SE} is the standard error of the EMM; in the \textit{network} column ER stands for Erd\H{o}s-R\'enyi and WS for Watts--Strogatz. Confidence intervals are Wald intervals from linear mixed-effects models and may extend beyond the theoretical $[0,1]$ range of directedness.}
\label{tab:agg-results}
\end{table}

\begin{table}[ht]
\centering
\begin{tabular}{lrccrc}
\toprule
 & $\operatorname{Cons}(0)$ & $\operatorname{Cons}(T)$ & $\Delta\operatorname{Cons}$ & \makecell{Runs at full \\ agreement, \(\%\)} & Runs \\
Scenario &  &  &  &  &  \\
\midrule
I. Baseline & 1.00 & 0.99 & -0.01 & 100.00 & 640 \\
II. Random Roles & 0.93 & 0.97 & 0.03 & 65.00 & 640 \\
III. Random Specialists & 0.70 & 0.88 & 0.18 & 0.00 & 640 \\
IV. Specialists (matched) & 0.72 & 0.86 & 0.15 & 0.00 & 640 \\
\bottomrule
\end{tabular}
\caption{\textbf{Mean initial and final consensus} by scenario ($n=640$ runs per scenario). 
\emph{Runs at full agreement}: share of runs starting at full consensus, 
$\Delta \operatorname{Cons}$: consensus change from initial to final round, \Cref{tab:agg-results}.}
\label{tab:consensus-levels}
\end{table}

\begin{table*}
\centering
\begin{tabular}{lccc|ccc|r}
\toprule
  & \multicolumn{3}{c}{Agent-level ICC (95\% CI)} & \multicolumn{3}{c}{Network-level ICC (95\% CI)} &  \\
Scenario & ICC & Lower & Upper & ICC & Lower & Upper & RLRT $p$ \\
\midrule
I. Baseline LLM & 0.00 & 0.00 & 0.36 & 0.40 & 0.18 & 0.67 & 1.00 \\
II. Random Roles & 0.81 & 0.40 & 1.17 & 0.69 & 0.34 & 1.11 & <0.001 \\
III. Random Specialists & 5.75 & 5.05 & 6.48 & 2.42 & 1.24 & 3.73 & <0.001 \\
IV. Specialists (Matched) & 2.95 & 2.41 & 3.48 & 2.25 & 1.19 & 3.60 & <0.001 \\
\cline{1-8}
\bottomrule
\end{tabular}
\caption{\textbf{Variance decomposition of outgoing influence by scenario}: intraclass correlations (ICC, \%) show the share of variance in outgoing influence that is attributed to agent identities (agent-level) or network realizations (network-level). Additionally, we provide the likelihood-ratio test results (RLRT). 
The agent component of outgoing influence is zero for base LLMs (I: \(0.00\%\)), small but non-zero in scenario II (\(0.81\%,\) \(p<.001\)), and largest with specialists (III: \(5.75\%\);  IV: \(2.95\%\), both \(p<.001\)). 
The variation in outgoing influence due to the LLM agents' identities (persistent opinion-leader-like agents) is higher when specialist LLMs are introduced; that is, in scenarios III and IV.
}
\label{tab:var-decomp}
\end{table*}


\begin{table*}
\centering
\begin{adjustbox}{width=\textwidth}
\begin{tabular}{llcccccccc}
\toprule
 &  & Cohen's & \makecell{\(95\%\) CI} & \makecell{\(95\%\) CI}  &  & &  &  &  \\
Outcome & Contrast & \(d\) & lower & upper & SE & \(t\)-stat & \(p\)-val & df & Sig  \\
\midrule
\multirow[c]{4}{*}{Plasticity} & Role effect & 0.34 & 0.08 & 0.59 & 0.014 & 2.67 & 0.010 & 59 & * \\
 & Specialization effect & 0.50 & 0.25 & 0.76 & 0.014 & 4.00 & <0.001 & 59 & * \\
 & Role--spec. alignment & -0.01 & -0.26 & 0.24 & 0.014 & -0.08 & 0.940 & 59 &  \\
 & Composition effect & 0.67 & 0.49 & 0.84 & 0.01 & 7.49 & <0.001 & 59 & * \\
\cline{1-10}
\multirow[c]{4}{*}{Directedness} & Role effect & -1.83 & -2.12 & -1.53 & 0.04 & -12.44 & <0.001 & 60 & * \\
 & Specialization effect & -0.58 & -0.88 & -0.29 & 0.04 & -3.96 & <0.001 & 60 & * \\
 & Role--spec. alignment & -0.01 & -0.31 & 0.28 & 0.04 & -0.08 & 0.937 & 60 &  \\
 & Composition effect & -1.50 & -1.71 & -1.29 & 0.028 & -14.46 & <0.001 & 60 & * \\
\cline{1-10}
\multirow[c]{4}{*}{\makecell{Outgoing \\ Influence}} & Role effect & 0.17 & 0.04 & 0.29 & 0.004 & 2.68 & 0.009 & 68 & * \\
 & Specialization effect & 0.19 & 0.07 & 0.32 & 0.004 & 3.09 & 0.003 & 68 & * \\
 & Role--spec. alignment & -0.08 & -0.21 & 0.04 & 0.004 & -1.32 & 0.191 & 68 &  \\
 & Composition effect & 0.23 & 0.15 & 0.32 & 0.003 & 5.33 & <0.001 & 68 & * \\
\cline{1-10}
\multirow[c]{4}{*}{Consensus Change} & Role effect & 0.41 & -0.03 & 0.85 & 0.022 & 1.88 & 0.066 & 59 &  \\
 & Specialization effect & 1.41 & 0.96 & 1.85 & 0.022 & 6.38 & <0.001 & 59 & * \\
 & Role--spec. alignment & -0.29 & -0.73 & 0.15 & 0.022 & -1.33 & 0.190 & 59 &  \\
 & Composition effect & 1.47 & 1.16 & 1.78 & 0.016 & 9.41 & <0.001 & 59 & * \\
\cline{1-10}
\bottomrule
\end{tabular}
\end{adjustbox}
\caption{\textbf{\textit{Marginal} effects for the four contrasts on agent- and population-level outcomes}~(\Cref{app:descriptive}).  
Each contrast is a single-degree-of-freedom comparison between scenarios, averaged across network structures: \textit{role effect} (II~vs.~I), \textit{specialization effect} (III~vs.~II), \textit{role--specialization alignment} (IV~vs.~III), and \textit{composition effect}
(\{III, IV\}~vs.~\{I, II\}).
Effect sizes are reported as Cohen's \(d\) with \(95\%\) confidence intervals; larger \(|d|\) indicates a larger standardized difference; the sign of \(d\) indicates the direction of the effect.
\texttt{SE} is the standard error of the unstandardized contrast \(c\), \(t\)-stat is the associated \(t\)-statistic, the \(p\)-value tests \(\text{H}_0\): \(d = 0\), \texttt{df} denotes Satterthwaite-approximated degrees of freedom, and \(*\) in the \texttt{Sig} column denotes \(p<0.05\).
Contrasts are computed from the contrast model with scenario-specific random intercepts for statement and network realization (\Cref{app:contrast-analysis}); SE is therefore larger than under the EMM model of \Cref{tab:agg-results}.
}
\label{tab:marginal-contrasts}
\end{table*}

\begin{table}
\centering
\begin{adjustbox}{ max totalheight=0.7\textheight}
\begin{tabular}{lclccccccccc}
\toprule
 &  &  & Cohen's & $d$ CI  & $d$ CI &  &  &  & $p$-val &  &  \\
Outcome & Network & Contrast & $d$ & lower & upper & SE & $t$-stat & $p$-val & Holm & df & Sig \\
\midrule
\multirow[c]{12}{*}{Plasticity} & \multirow[c]{4}{*}{ER} & Role effect & 0.40 & 0.14 & 0.65 & 0.014 & 3.09 & 0.003 &  & 63 & * \\
 &  & Specialization effect & 0.52 & 0.27 & 0.78 & 0.014 & 4.11 & <0.001 &  & 63 & * \\
 &  & Role--spec. alignment & -0.17 & -0.42 & 0.09 & 0.014 & -1.29 & 0.200 &  & 63 &  \\
 &  & Composition effect & 0.64 & 0.46 & 0.82 & 0.010 & 7.08 & <0.001 &  & 63 & * \\
\cline{2-12}
 & \multirow[c]{4}{*}{WS} & Role effect & 0.28 & 0.02 & 0.53 & 0.014 & 2.16 & 0.035 &  & 63 & * \\
 &  & Specialization effect & 0.48 & 0.23 & 0.74 & 0.014 & 3.77 & <0.001 &  & 63 & * \\
 &  & Role--spec. alignment & 0.15 & -0.11 & 0.40 & 0.014 & 1.14 & 0.257 &  & 63 &  \\
 &  & Composition effect & 0.69 & 0.51 & 0.87 & 0.010 & 7.66 & <0.001 &  & 63 & * \\
\cline{2-12}
 & \multirow[c]{4}{*}{Between} & Role effect & 0.12 & 0.03 & 0.21 & 0.005 & 2.59 & 0.011 & 0.112 & 90 &  \\
 &  & Specialization effect & 0.04 & -0.05 & 0.13 & 0.005 & 0.94 & 0.349 & 1.000 & 90 &  \\
 &  & Role--spec. alignment & -0.31 & -0.40 & -0.22 & 0.005 & -6.77 & <0.001 & <0.001 & 90 & * \\
 &  & Composition effect & -0.05 & -0.12 & 0.01 & 0.004 & -1.62 & 0.108 & 0.757 & 90 &  \\
\cline{1-12} \cline{2-12}
\multirow[c]{12}{*}{Directedness} & \multirow[c]{4}{*}{ER} & Role effect & -1.88 & -2.18 & -1.58 & 0.041 & -12.56 & <0.001 &  & 65 & * \\
 &  & Specialization effect & -0.61 & -0.91 & -0.31 & 0.041 & -4.06 & <0.001 &  & 65 & * \\
 &  & Role--spec. alignment & 0.14 & -0.16 & 0.44 & 0.041 & 0.94 & 0.352 &  & 65 &  \\
 &  & Composition effect & -1.48 & -1.69 & -1.27 & 0.029 & -13.95 & <0.001 &  & 65 & * \\
\cline{2-12}
 & \multirow[c]{4}{*}{WS} & Role effect & -1.78 & -2.07 & -1.48 & 0.041 & -11.85 & <0.001 &  & 65 & * \\
 &  & Specialization effect & -0.56 & -0.86 & -0.26 & 0.041 & -3.72 & <0.001 &  & 65 & * \\
 &  & Role--spec. alignment & -0.16 & -0.46 & 0.14 & 0.041 & -1.09 & 0.278 &  & 65 &  \\
 &  & Composition effect & -1.53 & -1.74 & -1.32 & 0.029 & -14.41 & <0.001 &  & 65 & * \\
\cline{2-12}
 & \multirow[c]{4}{*}{Between} & Role effect & -0.11 & -0.22 & 0.01 & 0.016 & -1.82 & 0.072 & 0.646 & 90 &  \\
 &  & Specialization effect & -0.05 & -0.17 & 0.07 & 0.016 & -0.87 & 0.387 & 1.000 & 90 &  \\
 &  & Role--spec. alignment & 0.30 & 0.19 & 0.42 & 0.016 & 5.21 & <0.001 & <0.001 & 90 & * \\
 &  & Composition effect & 0.05 & -0.03 & 0.13 & 0.011 & 1.17 & 0.245 & 1.000 & 90 &  \\
\cline{1-12} \cline{2-12}
\multirow[c]{12}{*}{Outgoing Influence} & \multirow[c]{4}{*}{ER} & Role effect & 0.18 & 0.04 & 0.31 & 0.005 & 2.62 & 0.010 &  & 87 & * \\
 &  & Specialization effect & 0.23 & 0.09 & 0.36 & 0.005 & 3.38 & 0.001 &  & 87 & * \\
 &  & Role--spec. alignment & -0.13 & -0.26 & 0.01 & 0.005 & -1.90 & 0.061 &  & 87 &  \\
 &  & Composition effect & 0.25 & 0.16 & 0.34 & 0.003 & 5.29 & <0.001 &  & 87 & * \\
\cline{2-12}
 & \multirow[c]{4}{*}{WS} & Role effect & 0.16 & 0.02 & 0.29 & 0.005 & 2.35 & 0.021 &  & 87 & * \\
 &  & Specialization effect & 0.16 & 0.02 & 0.29 & 0.005 & 2.35 & 0.021 &  & 87 & * \\
 &  & Role--spec. alignment & -0.04 & -0.17 & 0.10 & 0.005 & -0.55 & 0.585 &  & 87 &  \\
 &  & Composition effect & 0.22 & 0.12 & 0.31 & 0.003 & 4.59 & <0.001 &  & 87 & * \\
\cline{2-12}
 & \multirow[c]{4}{*}{Between} & Role effect & 0.02 & -0.08 & 0.12 & 0.003 & 0.37 & 0.711 & 1.000 & 90 &  \\
 &  & Specialization effect & 0.07 & -0.03 & 0.17 & 0.003 & 1.39 & 0.168 & 1.000 & 90 &  \\
 &  & Role--spec. alignment & -0.09 & -0.19 & 0.01 & 0.003 & -1.82 & 0.073 & 0.646 & 90 &  \\
 &  & Composition effect & 0.03 & -0.04 & 0.10 & 0.002 & 0.94 & 0.348 & 1.000 & 90 &  \\
\cline{1-12} \cline{2-12}
\multirow[c]{12}{*}{Consensus Change} & \multirow[c]{4}{*}{ER} & Role effect & 0.379 & -0.072 & 0.830 & 0.023 & 1.68 & 0.098 &  & 66 &  \\
 &  & Specialization effect & 1.531 & 1.080 & 1.982 & 0.023 & 6.78 & <0.001 &  & 66 & * \\
 &  & Role--spec. alignment & -0.265 & -0.716 & 0.186 & 0.023 & -1.17 & 0.245 &  & 66 &  \\
 &  & Composition effect & 1.588 & 1.269 & 1.907 & 0.016 & 9.94 & <0.001 &  & 66 & * \\
\cline{2-12}
 & \multirow[c]{4}{*}{WS} & Role effect & 0.448 & -0.003 & 0.899 & 0.023 & 1.98 & 0.052 &  & 66 &  \\
 &  & Specialization effect & 1.281 & 0.830 & 1.732 & 0.023 & 5.67 & <0.001 &  & 66 & * \\
 &  & Role--spec. alignment & -0.320 & -0.771 & 0.131 & 0.023 & -1.42 & 0.161 &  & 66 &  \\
 &  & Composition effect & 1.345 & 1.026 & 1.664 & 0.016 & 8.42 & <0.001 &  & 66 & * \\
\cline{2-12}
 & \multirow[c]{4}{*}{Between} & Role effect & -0.068 & -0.264 & 0.128 & 0.010 & -0.69 & 0.490 & 1.000 & 90 &  \\
 &  & Specialization effect & 0.250 & 0.054 & 0.446 & 0.010 & 2.53 & 0.013 & 0.067 & 90 &  \\
 &  & Role--spec. alignment & 0.055 & -0.141 & 0.251 & 0.010 & 0.56 & 0.576 & 1.000 & 90 &  \\
 &  & Composition effect & 0.243 & 0.105 & 0.382 & 0.007 & 3.49 & <0.001 & 0.005 & 90 & * \\
 \cline{1-12} \cline{2-12}
 \bottomrule
\end{tabular}
\end{adjustbox}
\caption{\textbf{\textit{Conditional} and \textit{interaction} effects of the four planned contrasts on agent- and population-level outcomes}~(Appendix~\ref{app:descriptive}).  Conditional effects are estimated separately for Erd\H{o}s--R\'{e}nyi (ER) and Watts--Strogatz (WS) networks. 
The \textit{Between} rows report interaction contrasts: the ER conditional contrast minus the WS conditional contrast. A non-significant \textit{Between} row indicates that the manipulation does not produce any detectable differences between the two network structures. The four contrasts are role assignment (II~vs.~I), specialization (III~vs.~II), role–specialization alignment (IV~vs.~III), and population composition 
(\{III, IV\}~vs.~\{I, II\}).
Effect sizes are reported as Cohen's \(d\) with \(95\%\) confidence intervals; larger \(|d|\) indicates a larger standardized difference; the sign of \(d\) indicates the direction of the effect.
\texttt{SE} is the standard error of the unstandardized contrast \(c\), \(t\)-stat is the associated \(t\)-statistic, the \(p\)-value tests \(\text{H}_0\): \(d = 0\), \texttt{df} denotes Satterthwaite-approximated degrees of freedom. For \textit{between} contrasts, the $p_{\mathrm{Holm}}$ reports Holm-adjusted \(p\)-values separately within the agent-level and run-level test families
(12 and 8 tests; the run-level family also includes the four interaction contrasts for final consensus, which are not shown here).
An asterisk in the \texttt{Sig} column indicates \(p<0.05\), using unadjusted p-values for conditional effects and Holm-adjusted \(p\)-values for interaction effects.
Contrasts are computed from the contrast model with scenario-specific random intercepts for statement and network realization (\Cref{app:contrast-analysis}); SE is therefore larger than under the EMM model of \Cref{tab:agg-results}.}
\label{tab:conditional-contrasts}
\end{table}


\begin{table}
\centering
\begin{tabular}{llcccc}
\toprule
 &  &  & \makecell{Final-state} & \makecell{\(95\%\) CI} & \makecell{\(95\%\) CI} \\
Scenario & Surrogate & Network & MCC (\%) & lower & upper \\
\midrule
\multirow[c]{8}{*}{I. Base LLMs} & \multirow[c]{2}{*}{M1} & ER & 96.8 & 96.0 & 97.7 \\
 &  & WS & 96.8 & 95.8 & 97.8 \\
\cline{2-6}
 & \multirow[c]{2}{*}{M2} & ER & 98.5 & 97.8 & 99.2 \\
 &  & WS & 98.3 & 97.6 & 98.9 \\
\cline{2-6}
 & \multirow[c]{2}{*}{M3} & ER & 98.1 & 97.4 & 98.9 \\
 &  & WS & 98.1 & 97.5 & 98.8 \\
\cline{2-6}
 & \multirow[c]{2}{*}{M4} & ER & 98.1 & 97.4 & 98.9 \\
 &  & WS & 98.1 & 97.5 & 98.8 \\
\cline{1-6} \cline{2-6}
\multirow[c]{8}{*}{II. Random roles} & \multirow[c]{2}{*}{M1} & ER & 89.9 & 88.4 & 91.4 \\
 &  & WS & 93.7 & 92.5 & 94.9 \\
\cline{2-6}
 & \multirow[c]{2}{*}{M2} & ER & 92.3 & 91.1 & 93.6 \\
 &  & WS & 94.3 & 93.3 & 95.3 \\
\cline{2-6}
 & \multirow[c]{2}{*}{M3} & ER & 92.9 & 91.9 & 93.9 \\
 &  & WS & 94.8 & 94.1 & 95.6 \\
\cline{2-6}
 & \multirow[c]{2}{*}{M4} & ER & 93.0 & 92.0 & 94.0 \\
 &  & WS & 95.0 & 94.2 & 95.7 \\
\cline{1-6} \cline{2-6}
\multirow[c]{8}{*}{III. Random specialists} & \multirow[c]{2}{*}{M1} & ER & 79.4 & 77.5 & 81.6 \\
 &  & WS & 85.1 & 82.8 & 87.1 \\
\cline{2-6}
 & \multirow[c]{2}{*}{M2} & ER & 81.1 & 79.1 & 83.3 \\
 &  & WS & 84.3 & 82.1 & 86.5 \\
\cline{2-6}
 & \multirow[c]{2}{*}{M3} & ER & 82.4 & 80.3 & 84.7 \\
 &  & WS & 86.5 & 84.3 & 88.5 \\
\cline{2-6}
 & \multirow[c]{2}{*}{M4} & ER & 84.6 & 83.2 & 86.3 \\
 &  & WS & 87.9 & 86.0 & 89.5 \\
\cline{1-6} \cline{2-6}
\multirow[c]{8}{*}{IV. Specialists (matched)} & \multirow[c]{2}{*}{M1} & ER & 83.1 & 81.5 & 84.7 \\
 &  & WS & 78.5 & 76.2 & 80.8 \\
\cline{2-6}
 & \multirow[c]{2}{*}{M2} & ER & 81.1 & 79.7 & 82.6 \\
 &  & WS & 79.1 & 77.0 & 81.4 \\
\cline{2-6}
 & \multirow[c]{2}{*}{M3} & ER & 82.8 & 81.3 & 84.5 \\
 &  & WS & 80.6 & 78.6 & 82.6 \\
\cline{2-6}
 & \multirow[c]{2}{*}{M4} & ER & 86.0 & 84.9 & 87.0 \\
 &  & WS & 82.2 & 80.5 & 83.8 \\
\cline{1-6} \cline{2-6}
\bottomrule
\end{tabular}
\caption{\textbf{Final-state Matthews Correlation Coefficient (final-state MCC) by scenario, surrogate model, and network structure}~(\Cref{app:dynamics}). We report MCC with \(95\%\) bootstrapped confidence intervals for one-step prediction of the final-round belief from the observed round-$(T-1)$ state, pooled over agents and held-out runs. Surrogate models are M1 (persistence), M2 (persistence + global belief composition), M3 (persistence + local neighborhood composition), and M4 (M3 + agent identity and social role); see~\Cref{tab:surrogate_models}. Higher MCC indicates better agreement with observed final-round beliefs. ER denotes Erd\H{o}s--R\'{e}nyi networks and WS denotes Watts--Strogatz networks.}
\label{tab:final-mcc}
\end{table}

\begin{table}
\centering
\begin{tabular}{llcccc}
\toprule
 &  &  & \makecell{Consensus} & \makecell{\(95\%\) CI} & \makecell{\(95\%\) CI} \\
Scenario & Surrogate & Network & fidelity (\%) & lower  & upper \\
\midrule
\multirow[c]{8}{*}{I. Base LLMs} & \multirow[c]{2}{*}{M1} & ER & 91.8 & 91.3 & 92.2 \\
 &  & WS & 92.3 & 91.9 & 92.8 \\
\cline{2-6}
 & \multirow[c]{2}{*}{M2} & ER & 98.4 & 98.2 & 98.6 \\
 &  & WS & 98.5 & 98.3 & 98.8 \\
\cline{2-6}
 & \multirow[c]{2}{*}{M3} & ER & 95.9 & 95.0 & 96.5 \\
 &  & WS & 95.2 & 94.2 & 96.0 \\
\cline{2-6}
 & \multirow[c]{2}{*}{M4} & ER & 95.9 & 95.0 & 96.4 \\
 &  & WS & 95.2 & 94.2 & 96.0 \\
\cline{1-6} \cline{2-6}
\multirow[c]{8}{*}{II. Random roles} & \multirow[c]{2}{*}{M1} & ER & 78.9 & 78.1 & 79.7 \\
 &  & WS & 82.1 & 81.2 & 82.9 \\
\cline{2-6}
 & \multirow[c]{2}{*}{M2} & ER & 95.1 & 94.7 & 95.5 \\
 &  & WS & 95.6 & 95.3 & 95.9 \\
\cline{2-6}
 & \multirow[c]{2}{*}{M3} & ER & 93.1 & 92.2 & 93.8 \\
 &  & WS & 93.4 & 92.6 & 94.1 \\
\cline{2-6}
 & \multirow[c]{2}{*}{M4} & ER & 92.9 & 92.0 & 93.6 \\
 &  & WS & 93.0 & 92.3 & 93.8 \\
\cline{1-6} \cline{2-6}
\multirow[c]{8}{*}{III. Random specialists} & \multirow[c]{2}{*}{M1} & ER & 73.4 & 72.6 & 74.3 \\
 &  & WS & 75.3 & 73.9 & 76.8 \\
\cline{2-6}
 & \multirow[c]{2}{*}{M2} & ER & 92.4 & 92.0 & 92.8 \\
 &  & WS & 92.2 & 91.5 & 92.8 \\
\cline{2-6}
 & \multirow[c]{2}{*}{M3} & ER & 91.0 & 89.7 & 92.1 \\
 &  & WS & 91.1 & 90.2 & 91.9 \\
\cline{2-6}
 & \multirow[c]{2}{*}{M4} & ER & 91.0 & 89.7 & 92.0 \\
 &  & WS & 89.8 & 88.5 & 90.9 \\
\cline{1-6} \cline{2-6}
\multirow[c]{8}{*}{IV. Specialists (matched)} & \multirow[c]{2}{*}{M1} & ER & 75.6 & 74.8 & 76.4 \\
 &  & WS & 75.9 & 74.7 & 77.1 \\
\cline{2-6}
 & \multirow[c]{2}{*}{M2} & ER & 92.9 & 92.6 & 93.3 \\
 &  & WS & 91.5 & 90.8 & 92.3 \\
\cline{2-6}
 & \multirow[c]{2}{*}{M3} & ER & 91.5 & 90.2 & 92.5 \\
 &  & WS & 89.3 & 88.2 & 90.4 \\
\cline{2-6}
 & \multirow[c]{2}{*}{M4} & ER & 91.7 & 90.4 & 92.7 \\
 &  & WS & 89.2 & 88.0 & 90.4 \\
\cline{1-6} \cline{2-6}
\bottomrule
\end{tabular}
\caption{\textbf{Consensus fidelity by scenario, surrogate model, and network structure.}
Consensus fidelity measures agreement between empirical and surrogate consensus trajectories, averaged over rounds and held-out runs (5-fold run-level cross-validation). We report mean fidelity with $95\%$ bootstrap confidence intervals obtained from \(1{,}000\) resamples of the 16 network realizations. Higher values indicate that a surrogate more closely reproduces the population-level agreement observed in \texttt{CoevolveSim}. Surrogate models are M1 (persistence), M2 (persistence + global belief composition), M3 (persistence + local neighborhood composition), and M4 (M3 + agent identity and social role); see~\Cref{tab:surrogate_models}. ER denotes Erd\H{o}s--R\'enyi networks and WS denotes Watts--Strogatz networks.}
\label{tab:consensus-fidelity}
\end{table}